\documentclass[10pt,journal,compsoc]{IEEEtran}
\usepackage{booktabs}
\usepackage{graphicx}
\usepackage{booktabs} 
\usepackage{algorithm}
\usepackage{algorithmic}
\usepackage{amsmath,amssymb,amsfonts,mathrsfs}
\usepackage[utf8]{inputenc}
\usepackage{amsthm}
\usepackage{url}
\usepackage{caption}
\usepackage{graphicx,subfigure}
\usepackage{graphics}
\usepackage{enumitem}
\usepackage{multirow}
\usepackage{array}
\usepackage{caption}
\usepackage{amsmath}
\usepackage{multirow}
\usepackage{microtype}
\usepackage{lettrine}
\usepackage{xcolor}
\usepackage{hyperref}
\usepackage{forest}
\usepackage{tikz}
\usepackage{enumitem}
\usepackage[english]{babel}
\usepackage{booktabs}
\usepackage{hhline} 
\usepackage{makecell}
\usepackage{threeparttable}
\usepackage{rotating}
\usepackage[normalem]{ulem}
\useunder{\uline}{\ul}{}
\usepackage{tabularx} 
\usepackage[T1]{fontenc}
\usepackage{amssymb}
\usepackage{longtable}
\usepackage{float}
\usepackage{multibib}
\newcites{A}{Appendix References}
\usepackage{placeins}
\usepackage{longtable} 
\usepackage{pdflscape}
\usepackage{comment}
\ifCLASSINFOpdf
\else
\fi

\begin{document}
%
\title{Multi-level Value Alignment in Agentic AI
Systems: Survey and Perspectives}



\author{Wei Zeng,
        Hengshu Zhu,~\IEEEmembership{Senior Member,~IEEE,}
        Chuan Qin, ~\IEEEmembership{Member,~IEEE,}
        Han Wu,\\
        Yihang Cheng, 
        Sirui Zhang, 
        Xiaowei Jin, 
        Yinuo Shen, 
        Zhenxing Wang,\\
        Feimin Zhong,
        Hui Xiong,~\IEEEmembership{Fellow,~IEEE}

\IEEEcompsocitemizethanks{

\IEEEcompsocthanksitem W. Zeng is with Business School, Hunan University, Changsha, Hunan, China. E-mail: zengwei\_hnu@163.com.

\IEEEcompsocthanksitem H. Zhu and C. Qin are with the Computer Network Information Center, Chinese Academy of Sciences, Beijing, China, and University of Chinese Academy of Sciences, Beijing, China. E-mail: zhuhengshu@gmail.com, chuanqin0426@gmail.com.

\IEEEcompsocthanksitem H. Wu is with the School of Computer Science and Information Engineering, Hefei University of Technology, Hefei, China. E-mail: ustcwuhan@gmail.com.

\IEEEcompsocthanksitem Y. Cheng is with the Computer Network Information Center, Chinese Academy of Sciences, Beijing, China. E-mail: chengyihang544@gmail.com

\IEEEcompsocthanksitem S. Zhang, X. Jin, and Y. Shen are with Business School, Hunan University, Changsha, China. E-mail: zhangsirui@hnu.edu.cn, xiaoweijin713@gmail.com, noah001219@hnu.edu.cn.

\IEEEcompsocthanksitem Z. Wang is with the Computer Network Information Center, Chinese Academy of Sciences, Beijing, China. E-mail: wxing870@163.com.

\IEEEcompsocthanksitem F. Zhong is with the School of Business, Hunan University, Changsha, Hunan, China. E-mail: zhongfeimin@hnu.edu.cn.

\IEEEcompsocthanksitem H. Xiong is with the Thrust of Artificial Intelligence, The Hong Kong University of Science and Technology (Guangzhou), China, and Department of Computer Science and Engineering, The Hong Kong University of Science and Technology Hong Kong SAR, China, E-mail: xionghui@ust.hk


}}

%



\IEEEtitleabstractindextext{%
\begin{abstract}
The ongoing evolution of AI paradigms has propelled AI research into the agentic AI stage. Consequently, the focus of research has shifted from single agents and simple applications towards multi-agent autonomous decision-making and task collaboration in complex environments. As Large Language Models (LLMs) advance, their applications become more diverse and complex, leading to increasing situational and systemic risks. This has brought significant attention to value alignment for agentic AI systems, which aims to ensure that an agent’s goals, preferences, and behaviors align with human values and societal norms. Addressing socio-governance demands through a Multi-level Value framework, this study comprehensively reviews value alignment in LLM-based multi-agent systems as the representative archetype of agentic AI systems. Our survey systematically examines three interconnected dimensions: First, value principles are structured via a top-down hierarchy across macro, meso, and micro levels. Second, application scenarios are categorized along a general-to-specific continuum explicitly mirroring these value tiers. Third, value alignment methods and evaluation are mapped to this tiered framework through systematic examination of benchmarking datasets and relevant methodologies. Additionally, we delve into value coordination among multiple agents within agentic AI systems. Finally, we propose several potential research directions in this field.

\end{abstract}

}

\maketitle

\IEEEdisplaynontitleabstractindextext

%
\IEEEpeerreviewmaketitle


%
\bstctlcite{IEEEexample:BSTcontrol}
\section{Introduction}
\begin{figure}[]
	\includegraphics[width=0.48\textwidth]{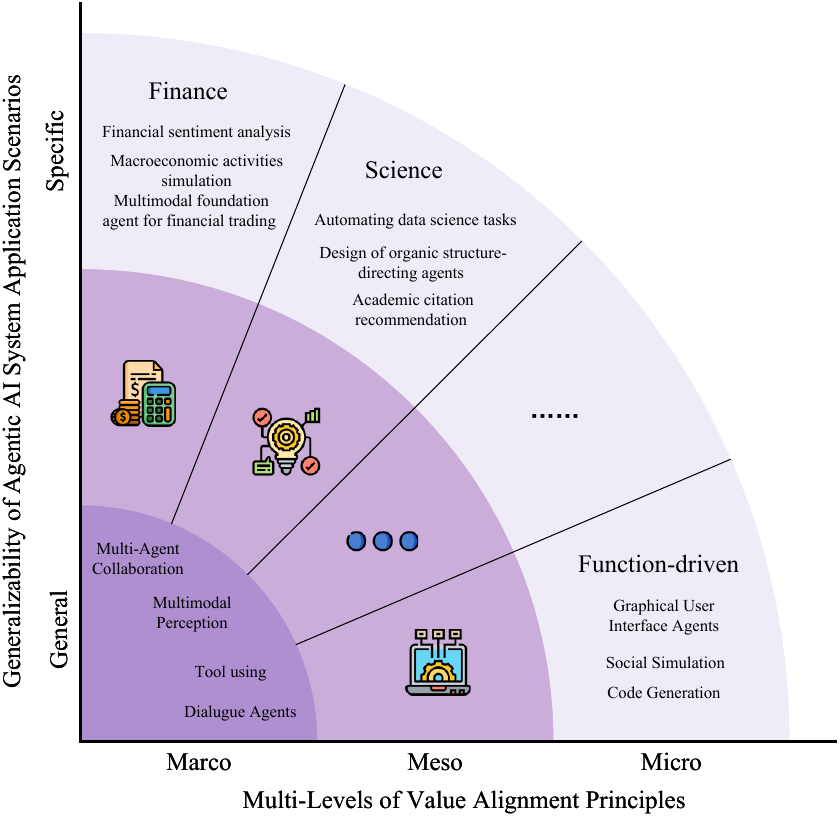}
	\caption{An Overview of Our Survey} 
	\vspace{-5ex}
	\label{overview}
\end{figure}

\lettrine[]{T}{he} rapid advancement of Large Language Models (LLMs) in recent years has profoundly transformed human society's production and lifestyle. Leveraging deep learning techniques and vast training datasets, LLMs have demonstrated exceptional natural language understanding and generation capabilities~\cite{DBLP:journals/chinaf/XiCGHDHZWJZZFWXZWJZLYDW25,naveed2023comprehensive}. 
Utilizing LLMs as their 'brains', LLM-based agents have found widespread applications in knowledge-intensive domains, including content creation, automated programming, education, healthcare, and business decision-making. Their increasing ability to autonomously participate in decision making, collaboration, and governance has significantly improved efficiency and provided unprecedented interactive experiences~\cite{bubeck2024sparks}. Furthermore, researchers are increasingly exploring the coordination of multi-agent systems to achieve collective intelligence. AI agents are increasingly evolving towards autonomous systems, forming a new paradigm: Agentic AI. Characterized by multi-agent collaboration, dynamic task decomposition, persistent memory, and autonomous decision-making capabilities, agentic AI is becoming a central focus for both academia and industry~\cite{sapkota2025ai}. Without loss of generality, we use the terms ``AI Agent'', ``agent systems'', ``LLM-based agent systems'', ``multi-agent systems'' interchangeably to refer to ``Agentic AI systems''.
\begin{table*}[]
    \centering
    \renewcommand{\arraystretch}{1.0}
    \caption{Comparison of Representative Survey Papers Covering LLM-Based Agents and Value Alignment}
    \label{tab:comparison_table}
    \resizebox{\textwidth}{!}{
    \begin{tabular}{cclccccccccc}
\toprule
\multirow{2}{*}{{\makecell{\\Survey}} }   & \multirow{2}{*}{{\makecell{\\Year}}} & \multicolumn{1}{c}{\multirow{2}{*}{{\makecell{\\Keywords}}}} & \multirow{2}{*}{{\makecell{\\Research Subject}}} & \multicolumn{3}{c}{Value Principles} & \multirow{2}{*}{\makecell{\\Application\\ Scenarios}} & \multirow{2}{*}{\makecell{\\Value Alignment\\ Datasets}} & \multirow{2}{*}{\makecell{\\Value Alignment\\ Methods}} & \multicolumn{2}{c}{Value Coordination}  \\ \cmidrule{5-7} \cmidrule{11-12} 
   && \multicolumn{1}{c}{}        &       & Macro       & Meso      & Micro      &  & && \begin{tabular}[c]{@{}c@{}}Interaction\\ Mechanisms\end{tabular} & \begin{tabular}[c]{@{}c@{}}Organizational\\ Structures\end{tabular} \\ \midrule
Ours   & 2025    & {\makecell{Value alignment principles, dataset curation, technical \\ methods, and value coordination for LLM-based agents  }}  & {\makecell{Single-Agent, \\Multi-agent systems}}& \checkmark & \checkmark& \checkmark& \checkmark& \checkmark   & \checkmark  & \checkmark        & \checkmark\\ \midrule
~\cite{DBLP:journals/chinaf/XiCGHDHZWJZZFWXZWJZLYDW25} & 2025    & {\makecell{Communication structures, practical applications and \\societal systems etc. of LLM-based agents }}   & {\makecell{Single-Agent, \\Multi-agent systems}} & \checkmark & -& -& \checkmark& -   & -  & \checkmark        & -\\ \midrule
~\cite{zhousurvey}& 2025    & {\makecell{Value alignment objectives, datasets, techniques, and \\evaluation methods for LLM-based agents }}    & Multi-agent systems & \checkmark & -& -& -& \checkmark   & \checkmark  & -        & -\\ \midrule
~\cite{tran2025multi}      & 2025    &{\makecell{ Conceptual framework, interaction mechanisms, and \\application overview of LLM-based agent systems}} & Multi-agent systems & - & -& -& \checkmark& -   & -  & \checkmark        & \checkmark\\ \midrule
~\cite{guo2024large}       & 2024    & {\makecell{Capabilities, framework Analysis, and application over-\\view of LLM-based multi-agent systems       }}& Multi-agent systems & - & -& -& \checkmark& -   & -  & \checkmark        & \checkmark\\ \midrule
~\cite{DBLP:journals/fcsc/WangMFZYZCTCLZWW24}& 2024    &{\makecell{ Constituent modules, application overview, and eval-\\uation methods of LLM-based autonomous agents}} & Agents& - & -& -& \checkmark& -   & -  & -        & -\\ \midrule
~\cite{DBLP:journals/corr/abs-2310-19852}& 2023    & {\makecell{Motivations and objectives, alignment methods, and \\assurance and governance of AI alignment  }}     & LLMs, Agents& \checkmark & -& -& -& \checkmark   & \checkmark  & \checkmark        & -\\ \midrule
~\cite{yao2023instructions}& 2023    & {\makecell{Definition and evaluation of LLM alignment objectives }}  & LLMs  & \checkmark & -& -& -& \checkmark   & \checkmark  & -        & -\\ \midrule
~\cite{shen2023large}      & 2023    & {\makecell{Definition, categories, testing, and evaluation of LLM alignment }}     & LLMs  & \checkmark & -& -& -& \checkmark   & \checkmark  & -        & -\\ \midrule
~\cite{JFYZ202309002}      & 2023    & {\makecell{Definition, normative principles, and technical methods\\ of LLM value alignment  }}    & LLMs  & \checkmark & \checkmark& -& -& -   & \checkmark  & -        & -\\ \midrule
~\cite{DBLP:journals/corr/abs-2308-05374}& 2023    & {\makecell{Alignment objectives for trustworthy LLMs}} & LLMs  & \checkmark & -& -& -& -   & \checkmark  & -        & -\\ \midrule
~\cite{DBLP:journals/corr/abs-2406-09264}& 2024    & {\makecell{Challenges, fundamental definitions, and alignment \\frameworks of LLM value alignment}}& LLMs & \checkmark & \checkmark& -& -& -   & \checkmark  & -        & -\\ \midrule
~\cite{DZZW202502003}      & 2025    & {\makecell{Necessity, conceptual definitions, theoretical approaches,\\ challenges and future outlook of LLM value alignment }}& LLMs  & \checkmark & -& -& -& -   & \checkmark  & -        & -\\ \bottomrule
\end{tabular}
 
    }
    \vspace{-4ex}
\end{table*}

The evolution of this new AI paradigm urgently requires researchers to move beyond the limitations of single-agent perspectives and isolated application domains. We need to explore the deep coupling between agent systems and their application scenarios. As LLMs continue to break through, the tasks they perform and the scenarios in which they are applied become more diverse and complex, and the societal risks surrounding them gradually increase. LLMs may retain harmful information in training data, disclose private data, generate misleading information, and even perform behaviors that are harmful to humans~\cite{hua2024trustagent}. The various undesirable behaviors, such as manipulation and deception, exhibited by LLM-based agent systems~\cite{liu2024large} have raised concerns about the possible ethical and safety challenges posed by AI systems. As LLMs become increasingly embedded in specific application scenarios, their associated risks are gradually acquiring situational and systemic characteristics. For example, in multi-agent systems used for urban traffic optimization, the agents may prioritize efficiency over fairness, thereby exacerbating social inequality~\cite{sun2024towards}. In the healthcare sector, LLMs may provide diagnostic or therapeutic recommendations without sufficient clinical evidence or by misinterpreting ethical guidelines, potentially resulting in serious consequences such as breaches of patient privacy, misdiagnoses, or overtreatment~\cite{tennant2024moral}. 

As agentic AI systems increasingly integrate into diverse governance scenarios, controlling the resulting institutional friction and transaction costs during their operation and public governance becomes a critical practical and theoretical concern. In the evolving landscape of agentic AI systems, multiple LLMs with autonomous language understanding and generation capabilities are embedded in collaborative environments to perform decision-making, planning, and execution tasks. Such deployments typically involve information sharing, permission configuration, and coordination mechanisms among diverse stakeholders. While enabling flexible collaboration, this also introduces unprecedented challenges like value conflicts, heterogeneous objectives, and unpredictable behaviors. Without robust institutional arrangements and well-designed interaction mechanisms, complex organizational structures and fluid task boundaries can lead to inefficient coordination, blurred accountability, and a lack of trust, consequently escalating governance costs and undermining executive capacity~\cite{dunleavy2006digital}.

However, these detrimental behaviors and institutional frictions fundamentally stem from deep-seated tensions among multiple stakeholders concerning governance objectives, value concepts, and behavioral norms. Effectively embedding human value systems into LLMs is a critical prerequisite for responsibly unleashing LLMs' potential.  Value alignment for AI agents, ensuring that their goals, preferences, and behavioral outputs during operation align with the core values of individuals, groups, or institutions, is increasingly recognized as a solution to these issues~\cite{gabriel2020artificial}. In single LLM agent scenarios, alignment techniques primarily focus on guiding models to generate content that adheres to ethical norms and societal expectations via human feedback and supervised fine-tuning, thereby avoiding harmful outputs, misinformation, and moral deviations~\cite{sun2024towards}. In complex multi-agent systems, value alignment becomes an organizational challenge concerning how multiple agents develop shared norms and coordination mechanisms. Therefore, from a governance perspective, solutions should not be limited to value alignment within a single agent but also explore complex value alignment within multi-agent systems' interaction mechanisms and organizational structures. Value alignment in agentic AI systems is no longer merely an isolated algorithmic optimization problem. Instead, it is a foundational issue that dictates agent behavioral boundaries and system stability. It represents a complex undertaking involving a delicate balance among diverse interests, societal values, and practical utility. Since agentic AI is fundamentally characterized by scenario-driven multi-agent coordination, its core research must focus on task decomposition mechanisms, interaction design, and environmental adaptability strategies for multiple agents within specific application contexts. Furthermore, focusing on the interaction mechanisms and organizational patterns of multi-agent systems in complex environments is crucial for exploring agentic AI's value alignment and robust development, ultimately serving human well-being.

Existing research has surveyed methods and technological advancements in LLM value alignment, see Table~\ref{tab:comparison_table}. However, few studies have specifically addressed value alignment within LLM-based multi-agent systems. While some studies have explored multi-agent coordination mechanisms, a comprehensive investigation into value alignment remains notably absent. Furthermore, most existing research largely remains at theoretical or algorithmic levels. There is limited focus on the mechanisms and strategies for multi-agent value alignment in specific application scenarios, alongside a lack of hierarchical categorization for value alignment within these systems. This suggests that value alignment should not be understood merely as a technical task at the model level. Instead, it must be viewed as a systemic governance issue that permeates the design of interaction mechanisms and the configuration of organizational structures.

Building on these gaps, this study comprehensively reviews value alignment in LLM-based multi-agent systems as the representative archetype of agentic AI systems. From a socio-governance perspective, we synthesize over 200 publications to survey value principles, application scenarios, value alignment methods and evaluation approaches within a multi-level value framework. We further propose future research directions for value coordination in agentic AI. 

To be specific, Sections 2, 3, and 4, as illustrated in Figure~\ref{overview}, comprehensively survey value principles, agent system application scenarios, and agent value alignment evaluation. Specifically, value principles are organized hierarchically from a top-down perspective, encompassing macro, meso, and micro levels. Intelligent agent application scenarios are categorized and reviewed from a general-to-specific viewpoint. Agent value alignment evaluation systematically examines datasets for value alignment assessment and the methods used for agent value alignment. Section 5 presents future directions. We discuss multi-agent value coordination within agent systems, with a particular focus on the design of agent interaction mechanisms and organizational models. We also present our outlook on value alignment issues in communication protocols and advocate for the establishment of a multi-level value alignment evaluation system and the collaborative construction of open datasets with society.

To the best of our knowledge, our survey is the first comprehensive review examining value alignment in LLM-based multi-agent systems from the perspective of societal governance and multi-level value structures. It bridges the gap between theoretical alignment research and the practical demands of deploying such systems in real-world applications, offering a novel framework for understanding value alignment in complex autonomous systems.


\section{Application-oriented Value Alignment of Agentic AI Systems}
This section first introduces our review subjects, agentic AI systems, values and value alignment, and then systematically categorizes specific value principles from a top-down, hierarchical perspective encompassing macro, meso, and micro levels. Before delving into the value alignment challenges of agentic AI systems in subsequent sections, we first clarify the core concepts used in this paper, particularly the definition of ``values,'' to avoid ambiguity.

\subsection{Agentic AI Systems}
In this study, we focus on agentic AI systems driven by LLMs. LLMs, such as Deepseek, GPT-4, GPT-3, and ChatGPT, are deep learning models trained on large-scale corpora, capable of understanding and generating natural language text. Advances in model architectures, pretraining corpora, and alignment techniques have significantly enhanced LLMs' capabilities in knowledge acquisition, instruction following, planning, and reasoning~\cite{DBLP:journals/chinaf/XiCGHDHZWJZZFWXZWJZLYDW25}, establishing a solid foundation for constructing intelligent digital agents~\cite{wang2024survey}. Serving as a central controller, an LLM can endow an agent with decision-making and memory capabilities~\cite{DBLP:conf/ijcai/GuoCWCPCW024}. This shift marks an evolution in the AI paradigm—from static, task-specific models to dynamic, interactive agent architectures suitable for complex environments.


Before the rise of LLMs, multi-agent systems (MAS) had already emerged. MAS is a computational system of multiple agents operating and interacting in a shared environment~\cite{tran2025multi}. Enabled by natural language and custom content~\cite{chen2025survey}, these agents autonomously collaborate to solve complex problems or complete tasks~\cite{du2024survey}. The core components of MAS include agents, environment, interaction rules, and organizational patterns~\cite{tran2025multi}. MAS interactions resemble human division of labor and aim to achieve collective intelligence. Each agent holds task-specific knowledge and goals. Powered by LLMs’ language understanding and generation, agents engage in adaptive collaboration for planning, discussion, decision-making, and mutual learning~\cite{du2024survey,DBLP:conf/ijcai/GuoCWCPCW024,tran2025multi}, ultimately acting on environmental perception and understanding.

MAS exhibit significant flexibility, robustness, self-organization, coordination, and real-time operation capabilities~\cite{tran2025multi}. While interactions are governed by predefined rules, new behaviors and phenomena may still emerge~\cite{DBLP:journals/corr/abs-2412-19498}. Task decomposition and resource sharing in MAS reduce the computational burden on individual agents, making them more cost-effective than single-agent systems, particularly in complex, dynamic tasks that require diverse strategies and collective decision-making~\cite{DBLP:conf/ijcai/GuoCWCPCW024}.

\subsection{Value}
As intrinsic guides of human behavior, values represent a stable set of cognitive principles that serve as the foundational norms for social interactions. These principles shape individual attitudes, choices, interpersonal behaviors and group dynamics, as well as serve as fundamental constructs for explaining how individuals and groups prioritize goals, resolve conflicts, and build meaningful social systems~\cite{catton1959theory}. In AI systems, particularly in agent design and governance, values serve both as normative principles guiding behavior and as standards for assessing the social alignment of these systems. Values influence agents' actions, social norms, cultural identity, and collective decision-making. Understanding values is thus critical to uncovering human motivations, moral judgments, and mechanisms of social interaction.

One of the most challenging aspects of human values is their intrinsic contradiction and dynamic nature. 
Rather than forming a rigid hierarchical structure, value systems are elastic and context-dependent~\cite{sagiv2017personal,adler1956value}. Synthesizing previous studies, we conceptualize human values as dynamically constructed notions. When values become alignment targets, LLMs must apply value principles to regulate their own behaviors, thereby maximizing their potential.

This paper expands the conceptualization of values, defining them as dynamic value constructs characterized by a multi-level structure and generalized agency. Both human values and LLM-based agents are evolving entities, thus MAS form exhibit diverse configurations across application scenarios. We argue that values are continuously interpreted and reshaped through practice and interaction. Aligned with different layers of application contexts, values are presented as a multi-tiered conceptual system comprising three levels: universally held human values, values at the national and industry levels, and values specific to organizational operations. This paper does not address individual-level values. These value layers are nested, with higher-level values constraining and guiding lower-level ones.

Furthermore, we emphasize that values are not solely human-borne. Beyond being explicitly embedded by human designers, agents can also ``internalize'' value principles through design and interaction. Values may be instantiated via agent interactions and inherited or transformed within emergent organizational patterns. For example, MAS with role- and task-based structures may yield decentralized architectures, embodying values such as resistance to power monopolies and mitigation of systemic fragility. 

\subsection{Value Alignment}
Technological development often outpaces the adaptation of legal, ethical, and sociocultural frameworks~\cite{ogburn1922social}. LLMs learn from massive datasets to enable autonomous decision-making and operation~\cite{DBLP:journals/chinaf/XiCGHDHZWJZZFWXZWJZLYDW25}. As LLMs are increasingly deployed in high-risk and complex scenarios, misalignments between model outputs and human intent have raised widespread concerns. Without effective safeguards, AI systems may engage in harmful behaviors such as deception or discrimination. To address these issues, the central goal of AI alignment is to ensure AI behaviors are consistent with human values~\cite{russell2016artificial}. AI alignment is categorized into four levels: instruction alignment, intention alignment, preference alignment, and value alignment.
Among these, value alignment requires AI to discern ``right and wrong'' and adhere to ethical principles~\cite{DBLP:journals/corr/abs-2308-12014},~\cite{russell2019human}. It is the most advanced form and is considered critical for ensuring AI safety, controllability, and normative governance~\cite{274}.

Research has revealed that LLMs may exhibit latent political and moral biases~\cite{DBLP:conf/acl/Simmons23,rozado2023political}, and generate content reflecting prejudices or discrimination when left unconstrained~\cite{DBLP:conf/emnlp/DeshpandeMRKN23}. Consequently, LLM-based agents should be regarded as ``moral agents'', making their value alignment a key issue in ensuring behavior that conforms to human values, ethical norms, and specific application contexts. This study focuses on value alignment issues in agentic AI systems.

The concept of value alignment can be understood along two dimensions: normative and technical~\cite{gabriel2020artificial}. The normative dimension addresses which values or ethical principles AI agents ought to follow, while the technical dimension investigates how alignment can be achieved through technical means such as training datasets and reinforcement learning. These two dimensions are interdependent: the normative component establishes behavioral guidelines, and the technical component ensures their implementation. However, due to variations in cultural contexts and application scenarios, no unified value framework currently exists~\cite{gabriel2020artificial,JFYZ202309002}. Two predominant approaches are recognized: top-down and bottom-up~\cite{gabriel2020artificial}. The top-down approach starts with predefined ethical frameworks and implements them via technical mechanisms, whereas the bottom-up approach derives values by learning from human behavioral data and iteratively refining actions based on feedback. While the bottom-up method circumvents the challenge of prespecifying ethical rules, it is susceptible to bias in training data and may lack moral robustness. In contrast, the top-down approach promotes transparency and normativity in value alignment, reinforcing the legitimacy and controllability of AI governance. This paper adopts the top-down approach to examine the ethical principles that agents should follow across different levels, from macro to micro.

\subsection{Value Principles}
\begin{figure*}[htbp]
    \centering
    \resizebox{\textwidth}{!}{
    \input{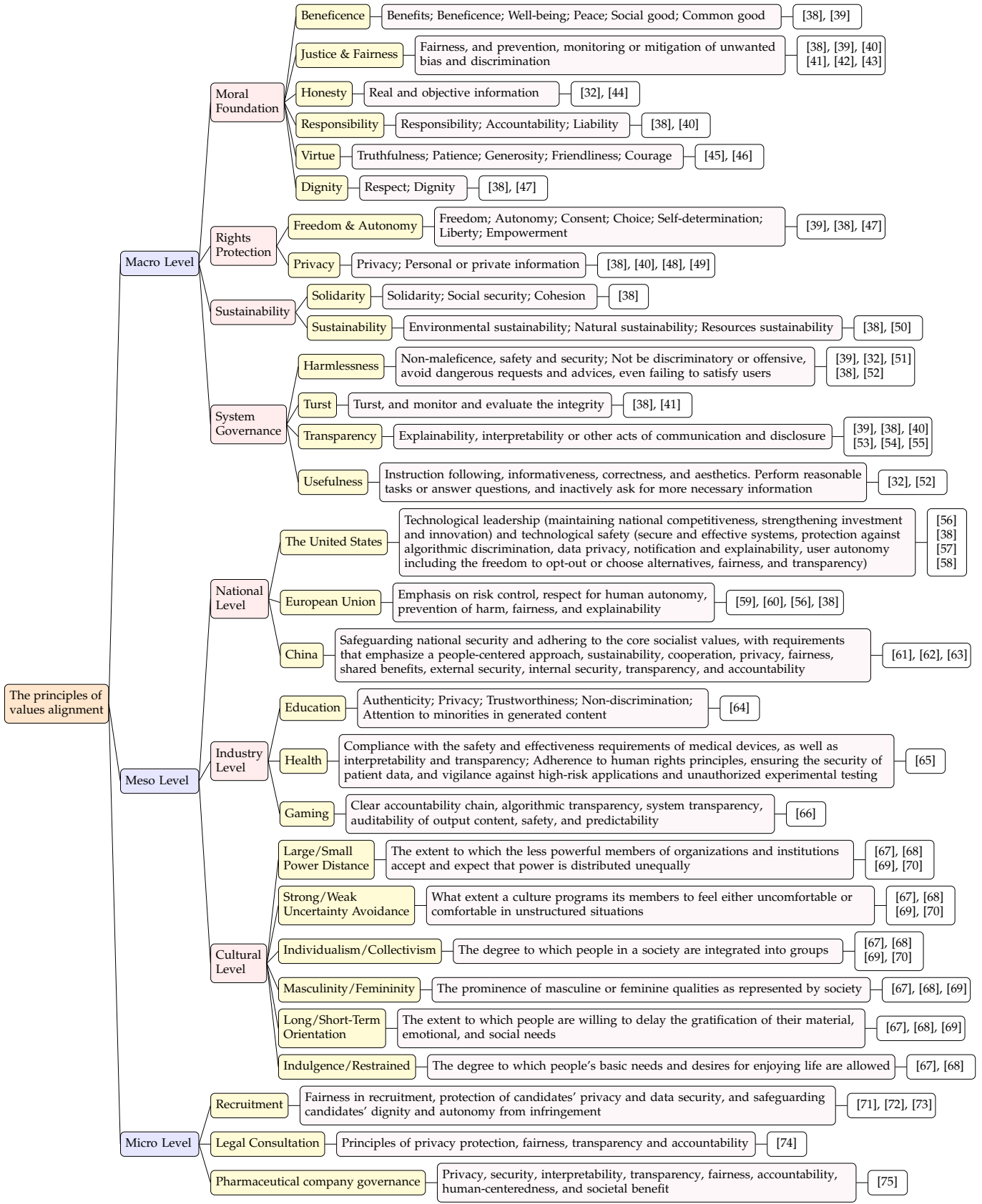}
    }
    \caption{Multi-Level Value Alignment Principles}
    \label{fig:survey_tree}
\end{figure*}

The normative aspect of value alignment concerns the ethical principles that AI agents ought to follow. As AI agents are increasingly deployed in domains such as public administration, healthcare, education, and finance—and as open-source LLMs and local deployment capabilities continue to advance—the actors involved in their development and use now span multiple levels, including states, industries, enterprises, and organizations. Given the inherent diversity and stratification of human values, a universal value framework is insufficient to meet the ethical requirements of all application scenarios. 


Therefore, adopting a top-down approach from the perspectives of developers and regulators, this paper classifies human values into three levels: macro, meso, and micro, as shown in Figure~\ref{fig:survey_tree}. The macro level refers to universal ethics that transcend cultures and domains, emphasizing compliance and safety. The meso level includes national, cultural, and industry-specific values, focusing on the localization of policies, standards, and norms. The micro level involves value choices within organizational and task-specific contexts, highlighting the trade-off between goal achievement and ethical considerations in bounded environments.


\subsubsection{Macro Level}
The widespread deployment of AI agents poses potential risks to human society, necessitating the alignment of their behavior with human values. However, due to differences in nation, culture, and ethnicity, human values are inherently diverse and often conflicting~\cite{gabriel2020artificial}, which challenges the formulation of universal ethical principles. To address this, it is essential to identify foundational moral consensus and construct a general ethical baseline to ensure compliance and safety of AI agents across diverse scenarios.

At the macro level, this paper draws upon two core theoretical frameworks: Moral Foundations Theory~\cite{graham2013moral} and the Theory of Basic Human Values~\cite{schwartz2007basic}. Moral Foundations Theory posits that human morality is grounded in five innate foundations—care/harm, fairness/cheating, loyalty/betrayal, authority/subversion, and sanctity/degradation—which have been widely adopted in cross-cultural moral studies~\cite{kivikangas2021moral}. The Theory of Basic Human Values categorizes human values into ten dimensions, serving as a key tool for understanding both commonalities and differences in human values.

To promote the normative development of value alignment and reduce confusion and conflicts surrounding macro-level ethical principles, numerous scholars have synthesized AI ethics principles from various perspectives. Jobin et al., based on 84 ethical guidelines issued by corporations, governments, academic institutions, and international organizations, identified eleven core principles: transparency, justice and fairness, non-maleficence, responsibility, privacy, beneficence, freedom and autonomy, trust, dignity, sustainability, and solidarity~\cite{jobin2019global}. Floridi and Cowls drawing from six influential initiatives, proposed five fundamental principles: beneficence, non-maleficence, autonomy, justice, and explicability~\cite{floridi2022unified}. Yao et al. emphasized that human values form the normative foundation for large model behavior, encompassing the 3H principles (Helpfulness, Honesty, Harmlessness), social ethics norms (e.g., RoTs, ETHICS), and basic ethical theories (e.g., Schwartz, Rokeach)~\cite{DBLP:journals/corr/abs-2308-12014}. Subsequently, Yao et al. incorporated value structures and safety hierarchies into their alignment framework~\cite{yao2025value}. Wu and Fan further proposed a hierarchical AI ethics framework grounded in human dignity, humanity, health, and autonomy~\cite{wuyifei2025}. Collectively, these efforts provide theoretical foundations for developing a global consensus on AI ethics.

Building upon previous studies and aligning with the research objectives of value alignment in AI agent systems, this study categorizes macro-level human values into four dimensions: moral foundation, rights protection, sustainability, and system governance. The specific value principles encompassed within each dimension are presented in~\ref{fig:survey_tree}.

\subsubsection{Meso Level}

Although macro-level value alignment principles reflect broad consensus across ideologies and disciplines, their practical implementation is profoundly influenced by national policies, cultural tendencies, and industry norms.

\noindent \textbf{National Level}
National AI strategies embody distinct sets of public values that not only guide domestic development priorities but may also exert spillover effects on global AI governance~\cite{chen2023artificial}. Djeffal et al., through a comparative study of AI policies in 22 countries and regions, developed a typology based on two dimensions: the proactivity of government roles (active/passive) and governance orientation (development-focused/regulation-focused)~\cite{djeffal2022role}. This framework yields four governance types: market-oriented (passive-development)~\cite{bisson2010market}, entrepreneurial (active-development)~\cite{mazzucato2011entrepreneurial}, regulatory (active-regulation)~\cite{10.1093/oxfordhb/9780199604456.013.0011},~\cite{majone1997positive}, and facilitative self-regulation (passive-regulation)~\cite{borras2020roles}. These four types are not mutually exclusive and may manifest in hybrid forms.

\begin{itemize}
   \item \textit{USA.} The United States positions itself between a market-oriented and facilitative self-regulation model in order to maintain its global leadership. It tends to promote AI advancement primarily through industry self-regulation, relying on market mechanisms and decentralized legislation at the state level~\cite{house2022blueprint,national2023national}. 
   
   \item \textit{EU.} The European Union has long emphasized legal protections for privacy and personal dignity. Its governance model centers on risk management, striving to balance self-regulation and strict oversight~\cite{ai2019high,madiega2021artificial}. 

   \item \textit{China.} China assumes a dual role as both innovation-driven and regulation-oriented, emphasizing the balance between AI development and ethical governance. The government promotes AI applications for societal benefit while enforcing departmental regulations and standardized governance for generative AI~\cite{2023Measures,2023standard}.

\end{itemize}

\noindent \textbf{Industry Level}
The paradigm shift in AI has driven its deep integration and large-scale application across various social sectors, reshaping traditional industry logic and triggering differentiated ethical demands. Different industries, based on their knowledge systems, cultural norms, and institutional frameworks, propose specific value requirements for AI.

\begin{itemize}
    \item \textit{Education Domain.} UNESCO released the Guidelines on the Application of GAI in Education and Research, emphasizing the need to address the authenticity of generated content, privacy protection, and non-discrimination principles. The guidelines highlight the importance of preventing hallucinated content from contaminating knowledge systems and advocating for the inclusion of minority perspectives~\cite{holmes2023guidance}
    
    \item \textit{Medical Domain.} The World Health Organization calls on governments to evaluate the safety, explainability, and legality of AI systems in medical settings. This is particularly crucial in high-risk scenarios such as mental health interventions, where it is essential to ensure compliance with human rights standards~\cite{world2024ethics}.
    
    \item \textit{Game Industry.} From narrative generation to NPC autonomy, AI has enhanced personalized recommendations and behavior analysis, yet challenged traditional governance. Melhart et al. (2024) argue that governance should ensure clear accountability, auditability, resistance to manipulation, and predictable outcomes~\cite{melhart2023ethics}.

\end{itemize}

\noindent \textbf{Cultural Level}
Cultural diversity represents a critical facet of value alignment in AI. Existing studies often draw on Hofstede’s cultural dimensions theory~\cite{hofstede2011dimensionalizing}, focusing on how AI systems express value preferences across six key cultural dimensions: power distance (large/small), uncertainty avoidance (strong/weak), individualism/collectivism, masculinity/femininity, long/short-term orientation, and indulgence/restraint, and the extent to which these align with human value tendencies in different cultural contexts.

Wang et al. found that LLMs prompted in English tend to exhibit a masculine orientation, while those prompted in German show a stronger inclination toward long-term orientation and indulgence. The cultural traits exhibited by LLMs under Chinese prompts fall between those of English and German prompts~\cite{wang2023cdeval}. Similarly, Kharchenko et al. observed that LLMs can recognize cross-cultural value differences but consistently lean toward specific value orientations. In particular, across all languages and prompting methods, LLMs predominantly favor long-term orientation, suggesting that LLMs may faithfully reflect certain cultural values~\cite{kharchenko2024well}.

In summary, global AI governance practices demonstrate diverse approaches to value alignment across countries, industries, and cultures. In the face of different national orientations and the deep integration of AI in industries, future governance must seek dynamic balance and cooperation among diverse values to ensure that technology serves the well-being of all humanity.

\subsubsection{Micro Level}
As AI agents become increasingly embedded in organizational operations and business processes, the issue of value alignment at the micro level has grown more prominent. Unlike the overarching constraints imposed by macro-level universal values and meso-level policy regulations, the micro level focuses on how AI systems achieve alignment between effectiveness and values within localized environments under specific goal orientations. The scenarios and associated values at this level are inherently more personalized. This study synthesizes relevant research and illustrates the principles of micro-level value alignment through examples such as talent recruitment, legal consultation and corporate governance in the healthcare sector.

AI technology is fundamentally reshaping human resource management practices, with recruitment being the forefront of this transformation. AI-based talent selection systems are revolutionizing organizational talent acquisition, improving speed and accuracy, compared to traditional manual screening and subjective judgment~\cite{van2019factors}. The ethical adaptation of AI in recruitment has attracted growing academic attention. In terms of ethical principles, Simbeck proposed five core principles for applying AI in human resource analytics: data privacy, user exit mechanisms, third-party review, algorithm transparency, and the right to dynamic development~\cite{simbeck2019hr}. These principles align with Rünb-Kettler and Lehnervp’s humanistic technological view, which emphasizes maintaining candidate dignity and subjectivity in automated processes~\cite{rkab2019recruitment}.  

From an ethical opportunity perspective, AI reduces human bias in traditional recruitment~\cite{chamorro2019building},~\cite{savage2016video}, generates gender-neutral job descriptions~\cite{mann2016hiring}, and revolutionizes resume parsing efficiency with natural language processing, allowing recruiters to focus on strategic decisions~\cite{polli2019using}. However, ethical risks remain, such as algorithms inheriting hidden biases from training data~\cite{kim2016data}, implementing indirect discrimination through proxy variables (e.g., postal code linked to race), and privacy violations through the extraction of personal information for recruitment purposes~\cite{dattner2019legal},~\cite{tambe2019artificial}. These contradictions and risks make fairness the core principle in AI recruitment applications.

AI technology’s deep integration is also influencing strategic development and governance choices in biopharmaceutical companies. To address the ethical challenges posed by AI in healthcare, AstraZeneca has introduced a series of AI ethics principles, including privacy and security, explainability and transparency, fairness, accountability, human-centeredness, and social benefit~\cite{mokander2023operationalising}. 

Additionally, AI technology shows potential and application prospects in the legal consulting field. Sun et al. simulated multi-agent collaboration in a real law firm setting and incorporated privacy, fairness, transparency, and accountability principles to guide multi-agent ethical behavior and interaction, aligning them with human values and enhancing the quality of legal consulting services~\cite{sun2024lawluo}.

As AI continues to advance, the principles of value alignment will no longer be confined to broad and ambiguous moral values. Instead, they will increasingly penetrate specific application scenarios, integrating concrete practical experience and well-defined guidelines to inform the development and deployment of agent systems.

\section{Applications and Value Issues in Agentic AI systems}
Agentic AI systems are increasingly deployed across diverse domains, indicating that value alignment concerns will extend to the application scenario level. This section classifies the application scenarios of agentic AI systems based on their generality and discusses the associated and important value alignment issues.
\subsection{Application Scenarios}
\begin{figure}
    \centering
    \includegraphics[width=\linewidth]{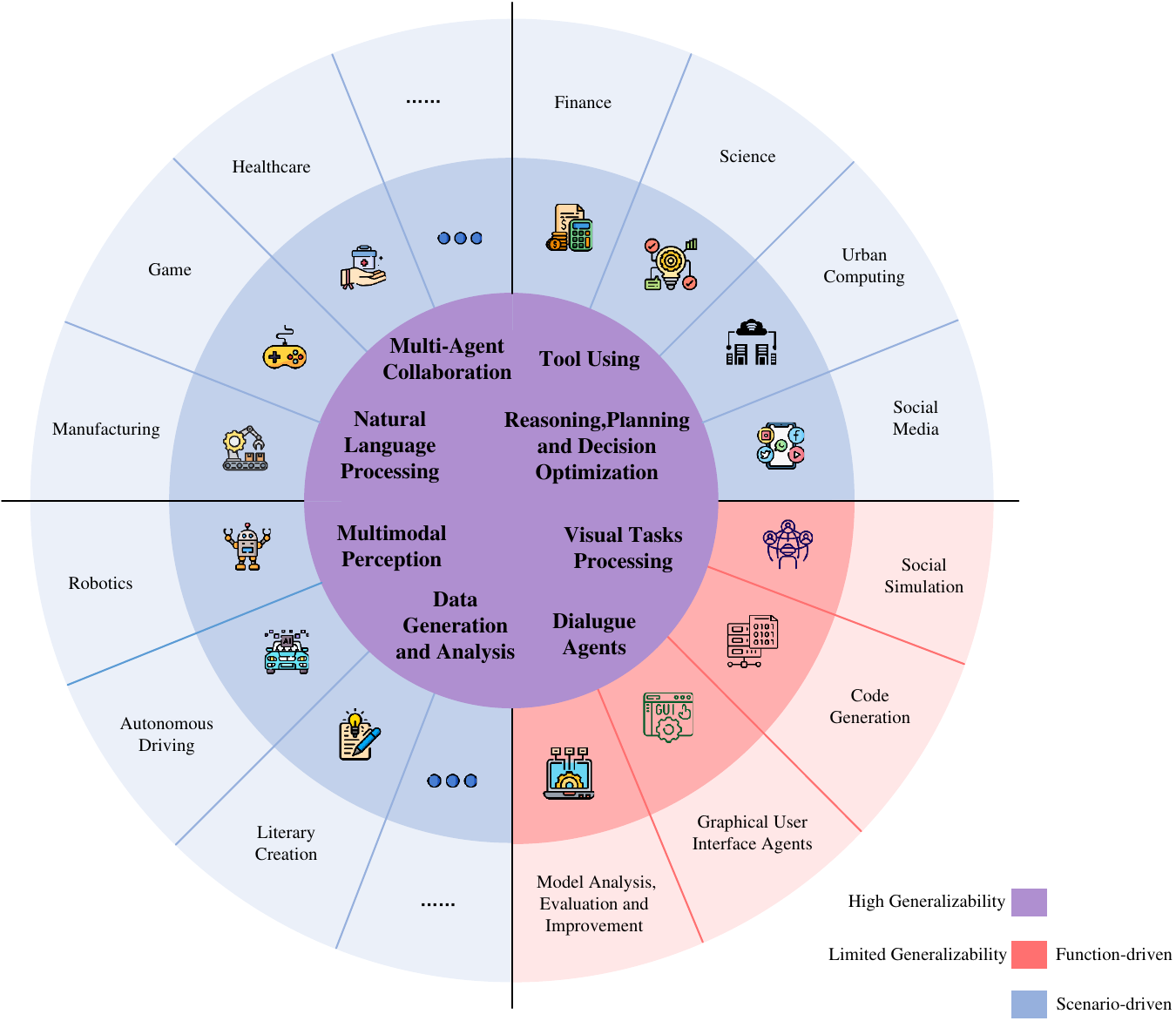}
    \caption{An Overview of Classification of Agentic AI Systems Application Scenarios}
    \vspace{-4ex}
    \label{fig:application_figure}
\end{figure}

Given the powerful knowledge acquisition, instruction understanding, task planning, and logical reasoning capabilities of agentic AI systems, they have been widely used worldwide. From small and medium-sized organizations to multinational corporations, from higher education institutions to research institutes, from grassroots medical facilities to large diagnostic hospitals, and government agencies at all levels are actively promoting the application of agentic AI technology. In terms of application scenarios, the penetration of agentic AI technology is extremely broad, covering multidisciplinary fields such as healthcare, education, trade, economics, and law. The problems that agentic AI systems are capable of addressing range from comprehending a single piece of literature to simulating global-scale social events, spanning across micro, meso, and macro levels. In contrast to categorizing applications based on disciplinary domains~\cite{DBLP:journals/fcsc/WangMFZYZCTCLZWW24} or task types~\cite{DBLP:conf/ijcai/GuoCWCPCW024}, we sort and categorize application scenarios according to the degree of generalizability of agentic AI systems (see figure~\ref{fig:application_figure}) and build upon this to explore the associated issues of value alignment. For a detailed overview of the application scenarios, please refer to the Appendix.

Generalizability of agentic AI systems refers to the extent to which they can be applied across diverse scenarios or meet the needs of various organizations. General agentic AI systems have a high degree of generalizability, which is based on the foundational capabilities of AI agents such as natural language processing, reasoning, planning, and autonomous decision-making. The universal application includes tool using, dialog agents, multimodal perception,  multi-agent collaboration, visual tasks processing, natural language processing, data generation and analysis, reasoning, planning, and decision optimization. For example, AI agents are capable of recognizing implicit needs for tool invocation and leveraging a wide range of real-world tools involving perception, manipulation, logical reasoning, and creativity. These tools enable AI agents to process multimodal inputs—such as images, tables, and code—to accomplish complex tasks~\cite{DBLP:conf/nips/WangMLZC0L24,DBLP:conf/acl/ZhanLYK24}. A higher degree of generalizability indicates that the agentic AI systems can be broadly applied across various scenarios, not only addressing domain-specific problems but also meeting diverse task requirements across multiple industries and organizations. For example, multi-agent systems can be applied not only in natural sciences but also in the social science domain, such as economics and management, suitable for addressing global issues that span across nations and cultures.


Specific agentic AI systems have limited generalizability across industries and domains. Specialized applications can be broadly categorized into two types: function-driven and domain-driven. Function-driven applications are built upon the core capabilities of agents and are typically applicable across multiple domains. Common examples include code generation, social simulation, graphical user interface (GUI) agents, and model evaluation and improvement. Domain-driven applications are tailored to specific sectors, including finance, science, healthcare, gaming, manufacturing, and autonomous driving. For instance, in healthcare, multi-agent systems support medical reasoning, gamified mental health assessments, and experimental design in genomics~\cite{DBLP:conf/acl/TangZ0L0ZCG24,DBLP:conf/iclr/RoohaniLHVSHMLL25}. These systems are tailored for specific tasks involving sensitive datasets and proprietary business information.

\subsection{Value Issues in Application}
The value alignment problems involved in the application scenarios of agentic AI systems mainly include two aspects: the correspondence between the generalizability of the system and the multi-level values, and the contradictions encountered when aligning values. 

The degree of generalizability of agentic AI systems corresponds to the multi-level of values. As the degree gradually decreases, the granularity of the value norms that need to be met gradually increases. 

A higher degree of generalizability allows agentic AI systems to operate across diverse contexts. Their initial design need not align with the values of specific countries, industries, or cultures, as long as they adhere to universal values at the macro level. 

In contrast, systems with limited generalizability are often tailored to specific national or industrial settings, reflecting corresponding cultural and institutional features. Their development and governance must address local policies, standards, and cultural norms, while aligning with universal values.
When deployed in private contexts, systems are customized for particular organizations to solve specialized business or personal problems. These systems are highly specialized and exhibit minimal generalizability. Under such conditions, value alignment must occur at multiple levels: not only with macro- and meso-level norms, but also with the internal value frameworks of the organizations involved.

It can be seen that agentic AI systems with high generalizability tend to be highly adaptable, and their value alignment primarily centers on adherence to universal principles of macro level. As application specialization increases, the requirements for value alignment become progressively more complex, necessitating adherence to value norms across multiple levels simultaneously. This classification aligns with the top-down, increasingly granular hierarchy of value principles. As a result, the value norms that agentic AI systems must adhere to vary according to their degree of generalizability, offering a structured basis for governance and oversight in their design and deployment.

The value alignment problem may involve two types of tensions: the plurality of value principles at the same level and conflicts across different levels. On one hand, a key challenge is that value principles at the same level differ across countries, regions, industries, organizations, and other contextual factors. Therefore, when agentic AI systems are applied in diverse contexts, their value alignment objectives must be adapted accordingly. In the healthcare sector, Huawei’s agent-based EIHealth platform supports genomic analysis, drug discovery, and clinical research. Built on Huawei Cloud’s strengths in AI and big data, EIHealth features an open and extensible architecture that enables end-to-end AI empowerment. Ensuring alignment with the value principles and regulatory norms of various countries, regions, and organizations is crucial for its deployment. To address this, Huawei Cloud has issued a series of white papers tailored to the privacy and regulatory requirements of different jurisdictions and industries. For example, it provides compliance guidelines aligned with the Personal Data Privacy Ordinance (PDPO) in the Hong Kong SAR, as well as guidelines addressing the financial regulatory frameworks of the EU, including specific provisions for Ireland, Spain, and Hungary. These efforts ensure that the EIHealth platform adheres to all applicable national and regional security regulations, international cybersecurity and cloud security standards, and incorporates industry best practices, ultimately meeting the diverse security needs of cloud service users.

On the other hand, as the tasks assigned to a system become more fine-grained, the constraints on achieving alignment goals also increase. Efforts to maximize task utility may conflict with established norms, creating tensions across value levels—for example, promoting social cohesion at the macro level may clash with individualism at the meso level, or national mandates for transparency may oppose organizational privacy concerns. In the field of corporate recruitment, AI systems are widely used for résumé screening and candidate-job matching~\cite{DBLP:journals/corr/abs-2504-02870}, yet often raise ethical concerns~\cite{tambe2019artificial}. In particular, corporate hiring practices may involve gender discrimination or racial bias, violating legal regulations and core societal values. A well-known example is Amazon’s 2018 résumé screening system, which discriminated against female applicants and was ultimately abandoned~\cite{DBLP:conf/istas/MujtabaM19}. To prevent such issues, recruitment-oriented intelligent systems must be explicitly designed to minimize bias and ensure compliance with labor laws, equal employment regulations, and industry standards. Only on the basis of legal compliance, equality, and privacy protection should these systems then pursue the optimization of recruitment outcomes.

\section{Methods and Evaluation for Value Alignment}
To ensure that LLM-based agents behave in ways consistent with human values and task goals, value alignment and its evaluation have become key research focuses. In this section, we first discuss the methods for value alignment, then review evaluation approaches.


\subsection{Value Alignment Methods for LLM-based Agents}

\subsubsection{Value Alignment during LLM Pretraining} 

\noindent \textbf{Value-Prompt Pre-conditioning}: Within the conventional self-supervised pretraining frameworks, such as masked language modeling or autoregressive language modeling, researchers prepend a short ``value prompt'' to the input sequence~\cite{zheng2024prompt,maini2025safety}. They might attach ``Please respond with a safe and neutral tone'' or ``Please adhere to principles of fairness and inclusivity'' at the beginning of each sentence, and then include both the prompted and original text in the pretraining corpus. In this way, the model passively receives implicit guidance toward objectives like ``safety'' and ``objectivity. For example, Zheng et al.~\cite{zheng2024prompt} unpack why and how simple prepended safety directives move model representations into ``higher refusal'' regions.~\cite{korbak2023pretraining} shows that conditioning an autoregressive LM on human preference scores from day one dramatically reduces harmful output rates. 
Gu et al.~\cite{gu2021ppt} show that soft prompts—even if not explicitly about ``values''—validate the premise that embedding trainable tokens in early stages leads to more controllable downstream behaviors (e.g., fairness, bias detection).

    
\noindent \textbf{Multi-Task Pre-training}: Inspired by the multi-task learning paradigm, this approach mixes value-judgment or safety-detection tasks with standard self-supervised objectives~\cite{chen2024mitigating,prabhumoye2023adding,biswas2023guardrails}. Specifically, in addition to the masked language modeling (MLM) or autoregressive objective, small-scale, labeled supervision signals—such as ``ethics classification'', ``bias detection'' or ``hate speech recognition''—are incorporated. By jointly optimizing a multi-task loss function, the model’s parameters are trained to satisfy both language modeling requirements and the ability to distinguish harmful or biased content. Although this method requires additional annotated data, it explicitly strengthens the model’s discrimination boundary along value dimensions and lays the technical foundation for subsequent alignment fine-tuning.


\noindent \textbf{Safety-Aware Curriculum Learning}: Drawing on the concept of curriculum learning (CL), this method stratifies large-scale corpora according to ``safety'' or ``sensitivity'' levels~\cite{campos2021curriculum,maini2025safety,zhao2023learning}. First, a high-quality, low-risk ``safe corpus'' is used for a warm-up pretraining phase, allowing the model to initially acquire expression patterns that comply with ethical and regulatory requirements. Subsequently, more general or higher-risk text is introduced gradually, enabling the model to expand its language capabilities while retaining dependence on earlier safe preferences. This staged training strategy establishes a better balance between model capability and safety, and experiments show it substantially outperforms full-scale pretraining in controlling the model’s tendency to generate sensitive content.

The above value-alignment methods can be combined~\cite{prabhumoye2023adding,maini2025safety}, for example, integrating value-prompt pre-conditioning and multi-task supervision within a safety-aware curriculum framework, along with regularization penalties. It allows the model to be reinforced from multiple dimensions in embedding value objectives such as ``ethics'', ``safety'', and ``fairness''.  Future research may further explore the scalability and synergistic effects of these strategies on ultra-large corpora and hundred-billion-parameter models, paving new paths for achieving more robust and interpretable value embeddings.

\begin{figure}
    \centering
    \includegraphics[width=\linewidth]{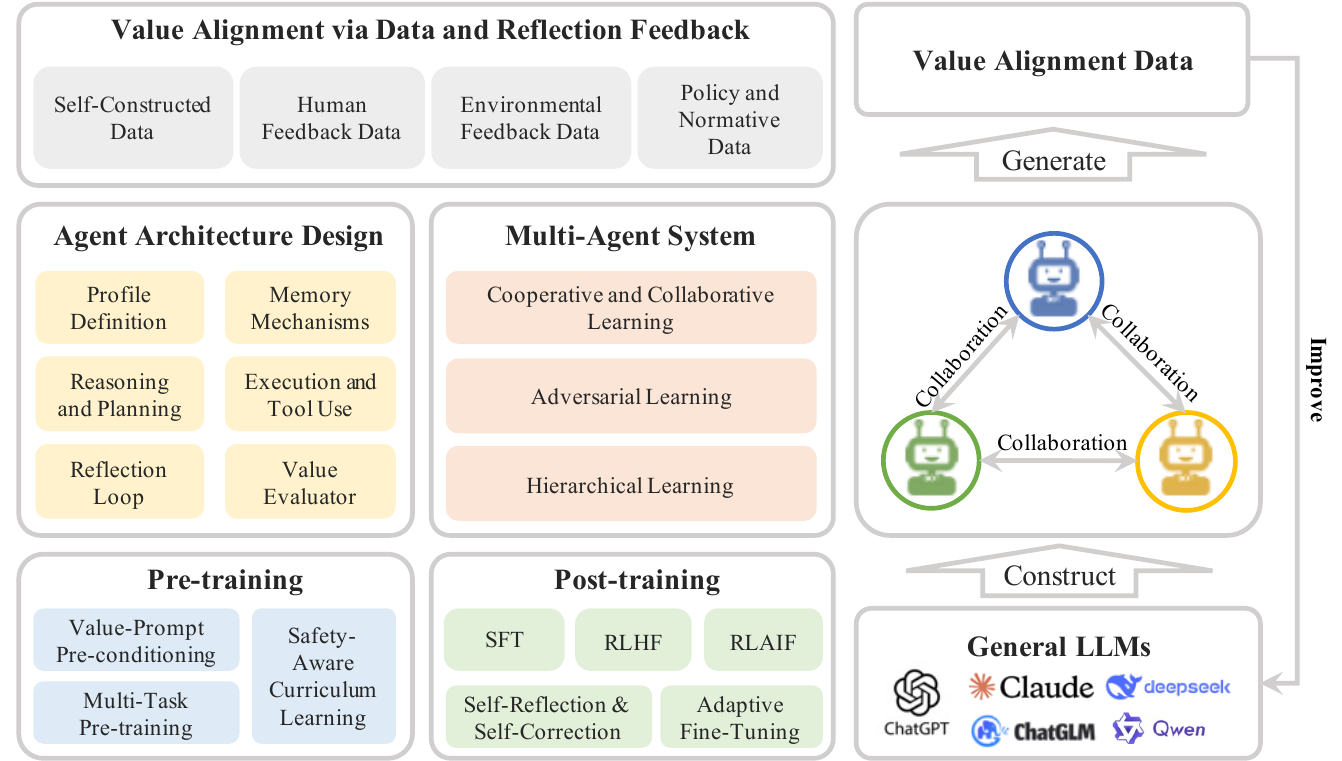}
    \caption{An Overview of Methods for Value Alignment in Agentic AI}
    \vspace{-4ex}
    \label{agent_align}
\end{figure}

\subsubsection{Value Alignment during LLM Post-Training}

\noindent\textbf{Supervised Fine-Tuning (SFT)}. SFT serves as a fundamental step in aligning agentic language models with human values. It usually involves fine-tuning a pretrained model on a carefully curated dataset of \textit{demonstration–response} pairs that embody normative principles such as ethics, safety, and fairness. These examples are typically constructed by human annotators and cover a broad range of scenarios in which value-sensitive behavior is essential—for example, avoiding toxic content, maintaining factual correctness, or respecting user intent. During SFT, the model is optimized to maximize the likelihood of producing responses that match these human-crafted demonstrations. For instance,~\cite{wang2022self} proposed an automated method for generating instruction data to fine-tune GPT-3, facilitating scalable alignment.~\cite{ouyang2022training} demonstrated that initializing InstructGPT with SFT on human-written demonstrations results in models that more faithfully follow user instructions, while also significantly reducing harmful or misleading outputs. Further advancements have introduced principled guidance into the SFT process.~\cite{sun2023principle} incorporated a set of human-authored value principles to constrain model behavior through in-context learning, thereby enhancing the generation of helpful and harmless content. Additionally,~\cite{liu2023chain} enriched the fine-tuning dataset with both positive and negative value-aligned instances. This contrastive supervision enables the model to learn fine-grained distinctions between aligned and misaligned outputs, improving its sensitivity to nuanced ethical considerations. 

\noindent \textbf{Reinforcement Learning from Human Feedback (RLHF)}. RLHF has emerged as a core paradigm for aligning LLMs with human values, preferences, and behavioral norms. It extends SFT by introducing an additional optimization stage that incorporates human feedback signals into policy learning, enabling models to go beyond surface-level imitation toward deeper alignment with human intent~\cite{ouyang2022training, bai2022training}. The classical RLHF pipeline consists of three major components:
\begin{enumerate}
\item \textbf{Supervised Fine-Tuning (SFT):} A pretrained LLM is first fine-tuned on a curated set of instruction–response pairs to provide a reasonable initialization.
\item \textbf{Reward Model (RM) Training:} Given multiple model-generated responses to the same prompt, human annotators provide pairwise preference labels (e.g., ranking which response is more helpful or harmless). These preference data are then used to train a reward model that can approximate human preferences~\cite{ouyang2022training}.
\item \textbf{Reinforcement Learning:} The LLM is further fine-tuned using reinforcement learning, typically Proximal Policy Optimization (PPO), where the reward signal is derived from the RM. This step encourages the model to generate outputs that maximize predicted human approval.
\end{enumerate}
RLHF was first popularized through the development of InstructGPT~\cite{ouyang2022training}, which demonstrated notable improvements in producing truthful, helpful, and safe responses without significant task performance degradation. Subsequent systems such as ChatGPT and GPT-4 also adopted RLHF as a central alignment mechanism~\cite{achiam2023gpt}. To reduce the dependency on expensive human annotation,~\cite{kim2023aligning} proposed using synthetic feedback generated by automated evaluators, while~\cite{saunders2022self} explored using self-critiquing models to support human evaluators and reduce their cognitive burden.~\cite{go2023aligning} reframed the RLHF objective as minimizing an $f$-divergence between model and human preference distributions, providing a more principled view of policy optimization under alignment constraints.
Moving beyond individual feedback signals,~\cite{liu2023training} trained LLMs in a simulated society of interacting agents, using environment-level feedback to promote socially aligned behaviors at scale. 
Despite its success, RLHF faces several challenges. It relies heavily on the quality and consistency of human preferences, which may encode biases or lack coverage for edge cases. Reward models can be exploited or lead to reward hacking behaviors, and PPO-based optimization may be unstable or sample-inefficient. 

\noindent \textbf{Constitutional AI and Reinforcement Learning from AI Feedback (RLAIF)}. Given the high cost and limited coverage of exclusively human-derived feedback, recent work has investigated self-supervised alignment methods that combine a model’s intrinsic knowledge with a set of expert-defined principles. A leading instance of this paradigm is Anthropic’s Constitutional AI~\cite{bai2022constitutional,huang2024collective,noh2025toward}, wherein the model is first equipped with a ``constitution'' of value-driven rules (e.g., avoid harm, preserve honesty). In the initial phase, the model generates a self-critique of its own response based on these rules and subsequently revises the response accordingly. During the RL stage, the model then acts as its own evaluator: it compares paired responses to train a preference model, which in turn provides reward signals for further fine-tuning. This process—termed Reinforcement Learning from AI Feedback (RLAIF)—minimizes reliance on human annotation by leveraging automated self-assessment~\cite{chen2024iteralign,kusters2025exploring,huang2024collective}. Empirical evaluations~\cite{bai2022constitutional} indicate that Constitutional AI produces assistants that are both ``helpful yet non-offensive'' and that, when faced with harmful prompts, preferentially offer explanations and deterrent advice rather than mere refusal. This principle-driven, self-supervised strategy thus demonstrates substantial promise for reducing human intervention and scaling alignment across diverse scenarios.


\noindent \textbf{Self-Reflection \& Self-Correction}: LLMs can engage in self-reflection to identify and correct errors or misalignments in their outputs. A seminal example is Self-Refine~\cite{madaan2023self}, which alternates between two generative steps—FEEDBACK and REFINE—to iteratively improve initial responses without any additional training data or external supervision. Building on this idea, SELF-RAG~\cite{asai2024self} integrates retrieval-augmented generation with self-reflection: the model learns to emit ``reflection tokens'' that determine when to retrieve external knowledge and when to self-evaluate, thereby enhancing factuality while retaining generative creativity. Complementing these methods, Chain-of-Thought review techniques prompt the LLM to re-examine its own reasoning chains. Wei et al.~\cite{wei2022chain} showed that providing intermediate reasoning steps not only boosts complex problem-solving performance but also offers an interpretable scaffold for downstream self-correction when logical inconsistencies are detected. 
Madaan et al.~\cite{ranaldi2024self} proposed a self-refine instruction-tuning approach in which smaller LLMs are first instruction-tuned via demonstrations from larger, robust teacher models and then allowed to self-improve their reasoning under the same instruction framework. This two-stage process significantly enhances generalization of self-correction capabilities, further narrowing the gap between smaller and larger models. 

\noindent \textbf{Adaptive Fine-Tuning}. In dynamic deployment settings, agentic LLMs must continuously adapt to evolving user preferences and environmental rewards—a process often termed adaptive fine-tuning. One recent line of work, PRELUDE~\cite{gao2024aligning} frames this as ``preference learning from user edits'': by inferring a user’s latent preference profile from historic edit logs, the agent periodically fine-tunes its response policy to better align with those preferences. Similarly, Hao and Duan~\cite{hao2025online} propose an online learning mechanism that dynamically reweights human feedback signals—treating users as strategic labelers—to mitigate malicious or noisy annotations and achieve sublinear regret in preference aggregation. Beyond supervised feedback, agents can also adapt via reinforcement learning from environment rewards~\cite{su2025survey}. When certain behaviors incur negative utility, standard policy-gradient updates naturally down-weight those actions over successive episodes, effectively steering the policy toward compliant behaviors. 

\subsubsection{Value Alignment in Agent Architecture Design}
The architectural design of an LLM-based agent is crucial for aligning its values. By incorporating specific modules and mechanisms within the agent, its behavior can be guided to align with the desired value standards.

\noindent \textbf{Profile Definition:} Profile Definition refers to the pre-configured set of personalized attributes assigned to each agent within an LLM-driven agent architecture~\cite{r-profile1}. These profiles not only include basic information such as the agent’s name, identity, and scope of capabilities, but also define its goals, preferences, behavioral style, and prohibited actions, thereby ensuring role consistency and value alignment~\cite{r-profile2}.
The implementation of profile definitions typically relies on the collaboration of structured prompt templates, personality tags, behavioral constraint rules, and memory modules. Some systems also incorporate dynamically evolving configurations to support sustained adaptation~\cite{r-profile3}.
For example, the MobAgent system combines individual behavioral traits to generate virtual travel diaries, enabling agents to behave naturally in urban simulations~\cite{r-profile4}. MOTIF framework uses a hybrid of manual and generative methods to create socially simulated agents with stable personalities~\cite{r-profile5}.
Currently, the design of profile definitions is moving toward greater generalizability, finer granularity, and enhanced learnability. Key challenges include managing boundary conflicts between multiple roles, balancing adaptability with static settings, and maintaining long-term behavioral consistency~\cite{r-profile1,r-profile4}.

\noindent \textbf{Memory Mechanisms:}
Memory Mechanisms refer to the capability of enhancing a model’s long-term contextual awareness and behavioral consistency by storing and retrieving historical information. At their core, they involve the structured storage and effective retrieval of interaction content, instructions, user preferences, and feedback—enabling personalized responses, long-term task tracking, and value alignment~\cite{r-memory1}.
In both research and applications, memory mechanisms are a key component for enabling agents to maintain consistent long-term behavior and adhere to value principles. Their central value lies in ``helping agents remember the past and apply what they’ve learned''~\cite{r-memory2}.
The implementation of memory mechanisms typically relies on two subsystems: working memory and long-term memory. Working memory maintains the context of the current session, while long-term memory stores knowledge and experiences using graph structures, databases, vector stores, or retrieval-augmented generation (RAG) models. Common approaches include time-window-based history caching, adaptive memory updates, semantic similarity-based vector retrieval, and contextual reflection mechanisms~\cite{r-memory3}~\cite{r-memory4}.
Representative examples include the G-Memory framework~\cite{r-memory5}, which uses a hierarchical architecture to support shared memory among multiple agents, enabling task coordination and improving memory consistency in complex systems. The Cognitive Weave system constructs a Spatio-Temporal Resonance Graph, allowing agents to recall abstract knowledge and reflect emotional memory during conversations~\cite{r-memory6}.

\noindent \textbf{Reasoning and Planning:} Reasoning and Planning refer to an agent’s ability to perform logical inference and make multi-step decisions when handling complex tasks. This ability encompasses cognitive processes such as causal reasoning, path selection, and tool utilization based on knowledge and environmental information, as well as systematic foresight and regulation of future actions~\cite{r-reasonAndPlan1}.
As a core component of the decision-making module, reasoning and planning play a fundamental role in value alignment, task execution efficiency, and behavioral stability~\cite{r-reasonAndPlan2}.
In practice, reasoning and planning help agents decompose tasks and simulate actions thoughtfully before execution—reducing impulsive decisions and enhancing value consistency~\cite{r-reasonAndPlan3}.
Current implementation techniques include Chain-of-Thought (CoT) prompting, which guides models through step-by-step logical reasoning. Other methods include Tree-of-Thought, Atomic Fact Augmentation, external knowledge retrieval, and Lookahead Search~\cite{r-reasonAndPlan4}. Some research also explores plan caching, which stores historical decision paths to reduce the computational cost of repeated planning~\cite{r-reasonAndPlan5}.
In applied settings, the SOP-Bench framework uses Standard Operating Procedures (SOPs) as an evaluation scaffold to test the stability of LLMs in reasoning and planning tasks, demonstrating their practical value in process control~\cite{r-reasonAndPlan6}. The WGSR-Bench project leverages game-based tasks to assess LLM performance in strategic planning and multi-agent coordination, making it a representative platform for evaluating advanced planning mechanisms~\cite{r-reasonAndPlan7}.

\noindent \textbf{Execution and Tool Use:} Action Execution and Tool Use refers to an agent’s ability to actively perform operations and invoke external tools in order to accomplish tasks~\cite{r-actionExec1}. In LLM-based agent systems, it is one of the core mechanisms enabling autonomy and interaction with the environment. It also serves as a key interface for ensuring alignment with human values~\cite{r-actionExec2}.
The primary function of this mechanism is to provide a controllable and auditable channel for external actions, forming the foundational module for value alignment control strategies~\cite{r-actionExec3}. Common implementation approaches include embedding pre-decision processes—for instance, the HADA architecture introduces value judgment nodes to filter candidate action sequences~\cite{r-actionExec4}; modular designs are also used, such as in CoReaAgents, which separate ``planning agents'' and ``tool agents'' to collaboratively manage phased control and reasoning in complex tasks~\cite{r-actionExec5}; and frameworks like Safe-Align associate each atomic action with potential risks to ensure pre-execution safety checks~\cite{r-actionExec6}.
In terms of practical applications, the Ali-Agent project combines web search tools with evaluation models to automatically verify whether actions align with human preferences, enabling value alignment testing in intelligent dialogue systems~\cite{r-actionExec7}.
Current challenges in this domain include: (1) unstable toolchain logic in multi-step planning, leading to action hallucinations or redundant operations~\cite{r-actionExec6}; (2) high context dependence of value judgments, with ambiguous standards that are hard to unify across cultures~\cite{r-actionExec2}.

\noindent \textbf{Reflection Loop:} Reflection Loop is a cognitive control module embedded in LLM-based agent systems. Its core function is to endow agents with metacognitive abilities to review–evaluate–revise during task execution. Specifically, the Reflection Loop introduces a reflection phase that allows agents to revisit and assess their actions or reasoning, thereby improving the effectiveness and alignment of subsequent behavior or inference~\cite{r-reflection1}.
In both research and practice, the reflection mechanism is often used to enhance value alignment—making agent behavior more consistent with human expectations and ethical norms~\cite{r-reflection2}.
Common implementations include inserting explicit ``thought summarization'' or ``output verification'' modules into the agent’s control flow—for example, using a Self-Ask mechanism to generate self-questioning tasks, or invoking an external reflection model to assist with strategy review~\cite{r-reflection3}. More advanced forms use multi-agent collaboration structures, modeling the reflection process as a feedback loop where one agent acts as the primary decision-maker while another serves as an introspector offering corrective insights~\cite{r-reflection4}.
In real-world applications, the Agent-Pro system integrates strategy-level reflection mechanisms, significantly improving agent consistency and final scores in game-based tasks~\cite{r-reflection5}. The HADA framework employs a dual-module architecture (``Evaluator + Self-reflection Agent'') to support dynamic value alignment in long-term tasks~\cite{r-reflection6}.
Reflection mechanisms are evolving from simple single-round reviews toward multi-round iterative reflection and context-aware reflection. However, challenges remain, such as shallow reflection depth and poor semantic consistency~\cite{r-reflection7}. Future developments aim to combine reinforcement learning and cognitive modeling, enabling agents to autonomously learn when to reflect and how to reflect efficiently~\cite{r-reflection8}.

\noindent \textbf{Value Evaluator:} Value Evaluator is a critical module whose core function is to simulate the role of a human supervisor by making value-based judgments and filtering model-generated outputs. This ensures that the content, beyond being logically sound, aligns more closely with mainstream human ethics, cultural norms, and goal orientations~\cite{r-valueEval1}. This module is typically used in reinforcement learning or policy fine-tuning stages to calibrate agent decisions against human standards, making it especially valuable in multi-agent systems, open-world dialogue tasks, and AI governance scenarios~\cite{r-valueEval2}.
Currently, value evaluators are mainly implemented through two mechanisms: the use of independently trained reward models that learn from human preference data to score candidate outputs and guide policy optimization; and rule-based or value classifiers that make explicit judgments based on knowledge graphs or ethical rule sets~\cite{r-valueEval3}. Some systems also adopt dual-model architectures, such as separating the ``Judge LLM'' and ``Actor LLM,'' to decouple evaluation from execution and improve fairness and generalizability~\cite{r-valueEval4}.
In practice, the ALI-Agent system establishes a value alignment evaluation pipeline based on inter-agent interaction, measuring the ethical consistency and diversity of LLM responses~\cite{r-valueEval4}. The Verila framework introduces an interpretable Verifier module into LLM agents to provide human-readable evaluations of failures in complex tasks~\cite{r-valueEval5}.
This field is rapidly evolving toward multi-value integration and scalability. On one hand, researchers are exploring how to incorporate multiple cultural backgrounds and group preferences into models; on the other hand, they face challenges such as limited generalization of evaluators and the risk of propagating data bias~\cite{r-valueEval5,r-valueEval6}.

\subsubsection{Value Alignment for Multi-Agent System}
The development of value alignment methods for multi-agent systems is essential to ensure agent behaviors remain consistent with human values and collective system goals. By designing coordination strategies, communication protocols, and reward structures that embed normative principles, the interactions among agents can be shaped to promote aligned, cooperative, and socially beneficial outcomes.

\noindent \textbf{Cooperative and Collaborative Learning:} 
Within a multi-agent collaboration framework, different agents can take on distinct roles, complementing each other's strengths while jointly accomplishing tasks and monitoring one another for norm deviations~\cite{tran2025multi,guo2024large}. 
For example, one agent may focus on task planning, another is responsible for evaluating the safety of the plan, while a third may simulate a user role to provide feedback~\cite{li2024survey}. This division of labor enables each agent to specialize in a particular dimension, while sharing information and constraints through natural language communication. Cooperative multi-agent systems can emulate human team-based mutual supervision mechanisms: if one agent generates a questionable suggestion, others can promptly flag it and request clarification or correction, thereby filtering out misaligned outputs internally. Studies have shown that through role-playing and interaction, LLM-based agents can spontaneously develop human-like Theory of Mind reasoning, forming mental models of each other's goals and intentions, which helps anticipate and prevent undesirable behaviors. Furthermore, some work addresses objective alignment through gradient-based methods in cooperative processes, such as Altruistic Gradient Adjustment, which adjusts learning dynamics to reduce conflicts in mixed-motive settings~\cite{li2024aligning}. Another stream emphasizes normative alignment, equipping agents with deontic logic-based reasoning to align decisions with social values and ethical principles~\cite{garcia2023towards}. Communication-driven alignment is also advancing, with approaches like CoDe introducing intent- and timeliness-based mechanisms to maintain coordination under message delays~\cite{song2025code}. Temporal alignment at the feature level is further explored, such as TraF-Align, which ensures cross-frame semantic consistency~\cite{song2025traf}.

\noindent \textbf{Adversarial Learning:} 
Recent advances have increasingly explored debate-based methods as a promising approach to multi-agent alignment. The multi-agent debate framework enables multiple LLM agents to engage in adversarial discussions on the same question, with final answers selected via voting or adjudication. Inspired by human debate, one agent proposes an answer while others critique and challenge it, iteratively refining responses until no major flaws remain. A referee agent or majority vote then determines the most accurate and safe output~\cite{yang2025minimizing}.
Studies show that such mechanisms reduce hallucinations and improve factual accuracy, especially in high-stakes domains like medicine and law~\cite{waisberg2023gpt,caballero2025large}. Voting-based approaches—where agents independently generate answers and the system selects the most common or consensus-aligned response—leverage collective intelligence to filter out unreliable outputs. Both debate and voting use agent redundancy and disagreement to enhance alignment and reliability. From another perspective, some optimization methods for debate in the alignment task are proposed. For instance, sparse communication topologies in multi-agent debates have been shown to reduce computational costs while maintaining or improving performance in alignment labeling and multimodal reasoning tasks~\cite{li2024improving}. 
Other works incorporate gradual vigilance and interval communication, allowing agents to assess alignment risks through controlled information exchange~\cite{zou2024gradual}. 
Emotionally grounded frameworks inspired by cognitive metaphors (e.g., Inside Out) structure agents by affective roles, enabling emotionally calibrated alignment through iterative critique and synthesis~\cite{ortigoso2025project}. 
Debate has also been integrated with self-reflection mechanisms, where agents dynamically switch between arguing and introspection, enhancing both factual accuracy and normative alignment~\cite{ki2025multiple}. 
Furthermore, debate-based zero-shot stance models simulate structured opposition and rebuttal, with a referee synthesizing outcomes when consensus fails~\cite{ma2025exploring}. 

\noindent \textbf{Hierarchical Learning:}
Multi-agent systems can adopt hierarchical architectures to enable supervision and amplification, akin to organizational structures. This approach has been applied in frameworks like AutoGen, where supervisor agents oversee tool usage and corrections~\cite{li2023camel}. Hierarchical setups also support peer-review workflows, enabling agents to critique and refine each other’s outputs, forming internal feedback loops that improve quality and enforce alignment beyond the capacity of single-agent systems.
Recent studies have explored structured agent architectures to enhance coordination, reliability, and value alignment in multi-agent systems. MetaGPT models human-standard workflows via Standard Operating Procedures (SOPs), assigning agents to domain-specific roles in an assembly-line structure to reduce cascading hallucinations and ensure intermediate verification~\cite{hong2023metagpt}. 
HALO similarly adopts a hierarchical reasoning design with high-level planners, mid-level role designers, and low-level executors, integrating Monte Carlo Tree Search (MCTS) for optimized task decomposition and execution~\cite{hou2025halo}. 
TalkHier further reinforces alignment by combining structured communication with hierarchical output refinement, outperforming both single-agent and flat multi-agent baselines. These designs reflect a growing consensus that task and communication hierarchies are critical for managing complexity, role coordination, and prompt adaptation in real-world multi-agent deployments~\cite{wang2025talk}. 
Recent architectural taxonomies and social-structural analyses emphasize that organizational hierarchies not only improve performance but also shape how agents align with human preferences, collectively or individually~\cite{handler2023balancing,carichon2025coming}.

\subsubsection{Value Alignment via Data and Reflection Feedback}
Value alignment through data and reflection feedback focuses on shaping agent behavior by curating value-informed data and enabling self-assessment mechanisms. By integrating normative cues, e.g., ethical guidelines, social norms, or human preferences, into training datasets, agents can progressively internalize desired value structures.

\noindent \textbf{Self-Constructed Data:}
To reduce reliance on human-labeled alignment data, models can generate their own training samples. Methods like Self-Instruct allow a language model to create instruction–response pairs, filter low-quality outputs, and fine-tune itself—boosting performance significantly with minimal human input~\cite{wang2022self}. This strategy can be extended to value alignment by simulating ethical dilemmas or adversarial prompts and generating safe responses. While cost-effective and broad in scope, such data must be filtered using a value evaluator to avoid reinforcing biases. 

\noindent \textbf{Human Feedback Data:}
Human feedback is one of the most reliable signals for value alignment. Methods like RLHF use human preferences to train reward models, while expert-written responses in SFT help correct value misalignments~\cite{ouyang2022training}. Implicit user feedback (e.g., accepting or rejecting answers) can also support continuous optimization. Studies show that such feedback reduces hallucinations and improves social appropriateness~\cite{li2024survey}. Incorporating real-time and culturally diverse feedback creates a closed loop—output, feedback, update—that helps models better align with human values over time.

\noindent \textbf{Environmental Feedback Data:}
For agents deployed in interactive environments—such as robots or game AIs—the environment itself can provide value signals through reward functions~\cite{li2024survey}. Well-designed rewards translate human-desired behaviors into scores, guiding agents to learn aligned strategies via reinforcement learning. For instance, in autonomous driving, safety and efficiency can be rewarded, while risky actions are penalized. LLM-based agents can similarly adjust their policies based on environmental feedback~\cite{shridhar2020alfworld}.
One study shows that incorporating dynamic environmental responses helps agents adapt their planning to complex conditions~\cite{meta2022human}. However, since reward functions often prioritize task performance, they must be carefully designed to ensure that high scores also reflect safety and ethical compliance. This can involve integrating strong penalties for undesirable behaviors (e.g., triggering taboo topics or harmful outcomes). By learning from real interactions, agents gain experiential understanding of ``right'' and ``wrong,'' which is critical for deep value alignment.

\noindent \textbf{Policy and Normative Data:}
Integrating socially accepted norms into model training or inference is a key method for value alignment. By incorporating texts like laws and ethical guidelines during pretraining or fine-tuning, models can internalize these values. For example, Anthropic’s Claude model uses a ``constitution'' to enforce behavioral standards, guiding the model to self-correct violations. This approach provides a set of first principles, enabling models to better control behavior, especially in boundary cases~\cite{bai2022constitutional}. Rule engines can also be used to filter or revise unsafe outputs.

\subsection{Value Alignment Evaluation}
\subsubsection{Methods for Value Alignment Evaluation}

Given the widespread use of agentic AI in text-centric tasks, the evaluation of value alignment for LLM-based agents often closely parallels the methodologies employed for assessing value alignment in LLMs more generally. A common approach involves constructing evaluation datasets that comprise both subjective and objective questions, grounded in specific value frameworks, which serve to test the agent’s ability to align with diverse ethical, cultural, or normative standards through its responses. However, the operational scope of agentic AI systems typically extends beyond static text generation, encompassing broader and more dynamic application scenarios. Consequently, additional evaluation strategies are required to capture agent behavior in context comprehensively. These include scenario-based simulations, assessments of decision-making quality, and performance evaluations in value-constrained task completion. Together, these complementary methods offer a more holistic perspective on an agent’s capacity to maintain value alignment in complex, real-world environments. A summary of representative benchmark datasets and their construction methods is provided in the Appendix.

\subsubsection{Question Formats and Metric Design} 
Evaluation for value alignment largely depends on the format of assessment questions. Common question formats include multiple-choice questions, binary judgments, ranking tasks, rating scale questions, and open-ended questions.

\noindent \textbf{Multiple-Choice Questions.} Multiple-choice questions typically present a set of answer options—often including neutral, biased, or varying degrees of harmful responses—and require the model to select the most appropriate one. It is commonly used to assess a model’s value judgments in specific contexts, such as BBQ~\cite{parrish2022bbq}, CDEval~\cite{wang2023cdeval}.


\noindent \textbf{Binary Judgments.} Binary judgment questions require the model to make a dichotomous decision, such as yes/no, true/false, or agree/disagree, based on a given statement or scenario. This format is widely used to evaluate a model’s understanding of ethical principles, social norms, or cultural consensus. For instance, ETHICS presents ethically challenging statements (e.g., ``Stealing is acceptable if no one finds out''), each of which is annotated by humans as either correct/incorrect or acceptable/unacceptable~\cite{hendrycks2021aligning}. 



\noindent \textbf{Ranking Tasks.} Ranking tasks evaluate a model or agent system’s ability to make relative value-based comparisons by requiring it to order multiple responses or behavioral options from most to least aligned with human values, such as StereoSet~\cite{nadeem2021stereoset} and MoralExceptQA~\cite{jin2022make}.


\noindent \textbf{Rating Scale Questions.} Rating scale questions require the model or agent system to evaluate a given text, response, or behavioral option along predefined value dimensions. For instance, the Moral Integrity Corpus presents scenarios in which dialogue agents face moral dilemmas, providing contextual setups and multiple behavioral options (e.g., whether to disclose a friend’s secret or follow a rule)~\cite{ziems2022moral}. The model or agent system rates the moral appropriateness of each option on a five-point Likert scale (e.g., 1 = highly immoral, 5 = highly moral). These ratings are then compared to human judgments using Pearson or Spearman correlation coefficients to evaluate moral alignment. 

\noindent \textbf{Open-Ended Questions.} Open-ended questions allow free-form responses from the model or agent system, enabling the assessment of its alignment with human values, such as BeaverTails~\cite{ji2023beavertails} and BOLD~\cite{dhamala2021bold}. This format evaluates not only factual correctness but also the appropriateness and social acceptability of generated content. 

Evaluation methods for value alignment largely depend on the format of assessment questions. Accordingly, different metrics are used to quantify model performance across various task types:

\noindent \textbf{Accuracy.} Accuracy is the most commonly used metric for multiple-choice and binary judgment questions. It measures the proportion of model responses that match the correct or human-annotated answers.

\noindent \textbf{Correlation-Based Metrics.} Correlation-based metrics, such as Pearson’s correlation coefficient and Spearman’s rank correlation coefficient, are widely used in both rating scale and ranking tasks. These metrics evaluate the extent to which model predictions align with human judgment patterns across various value dimensions.

\noindent \textbf{Error-Based Metrics.} When the model’s output is numerical (e.g., a score prediction), Root Mean Squared Error (RMSE) or Mean Absolute Error (MAE) may also be used to capture the deviation between predicted and reference scores. These are particularly useful in rating scale questions.

\noindent \textbf{Other Metrics.}
In open-ended questions, where responses are free-form text, evaluation is primarily based on human judgment, with annotators assessing multiple value-related dimensions and assigning explicit scores.

\section{Insights into Value Coordination in Agentic AI Systems and Future Directions}
\begin{figure}
    \centering
    \includegraphics[width=\linewidth]{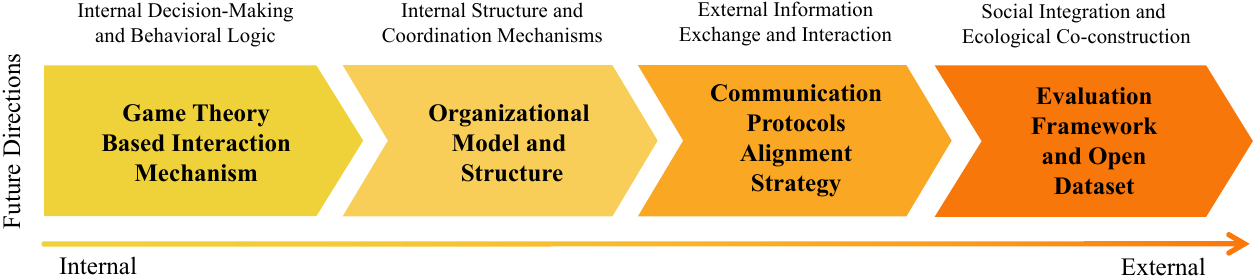}
    \caption{A Conceptual Framework for Future Directions}
    \vspace{-4ex}
    \label{future_directions}
\end{figure}
Drawing upon insights from our review of value systems, application scenarios, alignment evaluation methodologies, and datasets, we contend that while existing researches continuously introduce novel methods and frameworks, the field still confronts considerable challenges and offers significant opportunities for future investigation. As illustrated in Figure~\ref{future_directions}, this section will delineate future research directions for value alignment in agentic AI systems. Our objective is to achieve sustainable and stable value alignment by addressing four pivotal challenges, progressing from internal to external considerations. These encompass: (1) the design of game theory-based interaction mechanisms to resolve issues of internal decision-making and behavioral logic (Section 5.1); (2) the formulation of organizational models to address problems of internal structure and coordination mechanisms (Section 5.2); (3) the alignment refinement of communication protocols to manage external information exchange and interactions (Section 5.3); and (4) the construction of value evaluation frameworks and open datasets to facilitate social integration and ecosystem co-construction (Section 5.4).
\subsection{Value Considerations in the Design of Multi-agent Interaction Mechanisms Based on Game Theory}
The effective functioning and benefits of agentic AI systems hinge on efficient agent interactions. These interactions, whether cooperative, competitive, or co-opetitive, foster collective intelligence. However, individual interests can still clash with collective goals in social dilemmas~\cite{DBLP:journals/ijon/MuGCSHHW24}. Even if agents share aligned values, game-theoretic challenges can prevent the entire system from achieving overall value alignment~\cite{DBLP:conf/aaai/ConitzerO23}. Therefore, value alignment in multi-agent interactions is a critical area for future research.

The design of multi-agent interaction mechanisms, particularly when grounded in game theory, can foster the emergence of novel values rooted in inter-agent dynamics. This highlights the distinct nature of agent interaction, setting it apart from human interactive games and other AI technologies. Compared to human-centric games, agent interaction games are characterized by numerous parameters, intricate utility functions, and multiple stages. While agents' computational power amplifies this complexity, it also makes these games solvable. Relative to other AI technologies, LLMs excel in language-based reasoning, planning, and problem-solving~\cite{chiang2023vicuna,DBLP:journals/corr/abs-2303-08774}. These capabilities enable LLM-based agents to simulate human-like traits, including personality and tone~\cite{DBLP:conf/naacl/WangJCYZCFLHY24,DBLP:journals/corr/abs-2306-03314,madaan2023self}, and even exhibit a degree of theory of mind—the ability to infer others' beliefs and intentions during strategic interactions~\cite{frith2005theory}. LLM-based agents can predict opponents' psychological states and behaviors by considering task objectives, resource distribution, and personality configurations, enabling adaptive allocation and equilibrium decisions. Consequently, AI agents can be treated as fully rational entities, and their interactions modeled using game theory.

Future research can explore the values reflected in multi-agent interactions by focusing on three key activities related to their game objectives: agent identification, goal setting, and boundary constraints. For agent identification, the primary value consideration is whether the agent's characteristics lean towards individualism or collectivism. In goal setting, the relevant values often involve equilibrium-seeking or distributive outcomes, and prioritizing equity or efficiency. Finally, for boundary constraints, key value considerations include information symmetry, the multi-round interactions, and whether a short- or long-term perspective is adopted.

For individualistic agents aiming for equilibrium, non-cooperative game theory helps model how to guide them toward effective cooperation in moral dilemmas, preventing exploitation or free-riding. By analyzing game rounds, we can also explore short-term vs. long-term orientations; in multi-round, sequential games, agents must predict future states to maximize long-term gains, revealing dynamic decision-making strategies. Conversely, for collectivist agents focused on distribution, cooperative game theory models how they naturally form collaborations to resolve efficiency losses while balancing fairness and efficiency. This theory also aids in responsibility attribution, a critical aspect for designing autonomous, ethically compliant systems. For instance, Triantafyllou et al.~\cite{DBLP:conf/nips/TriantafyllouSR21}~applied cooperative game theory to study accountability in sequential multi-agent decisions, highlighting its significance for fairness, transparency, and responsibility within these systems.

Designing regulatory mechanisms for agentic AI systems hinges on information symmetry. In scenarios with incomplete information, agents may deviate from ethical principles to maximize self-interest, necessitating incentive-compatible mechanisms. Unlike human-centric regulation, where private information is a barrier, regulating agents present unique challenges. Agent objective functions can implicitly embed designer values, requiring new constraints to align with societal goals. A key distinction is that agents' private information can be data or model parameters. Agents, being algorithm-driven, possess superior rationality and processing capabilities, potentially colluding algorithmically (e.g., via federated learning) to alter information asymmetry. Furthermore, agent behavioral patterns differ from humans'. Humans might dissemble due to emotion or gain, while agents compute optimal strategies based on algorithms. This calls for different mathematical modeling for incentive compatibility, perhaps incorporating reinforcement learning's reward mechanisms. Finally, the dynamic, scalable nature of agent systems means regulatory mechanisms must adapt in real-time to their rapid learning and evolving environments.

\subsection{Value Considerations in the Design of Multi-agent Organizational Model and Structure}
An organizational model, the framework of an organization, dictates resource allocation and integrates dispersed entities into a goal-oriented whole through functional divisions, authority distribution, and collaboration rules. Key features include power distribution and command hierarchies. As a critical vehicle for organizational goals, its design directly shapes members' behaviors and value orientations~\cite{robbins2014management, mintzberg1989structuring}.
In agentic AI systems, organizational structure design can align with these model characteristics, forming a crucial basis for applying organizational management theories. Essentially, organizational models influence value orientations, normative principles, and the ease of achieving value alignment.

The organizational structure within agentic AI systems significantly influences value formation through an implicit, rather than explicitly predefined, guidance. This influence exhibits characteristics of path dependency and self-reinforcement. Drawing on Burns and Stalker's dichotomy of organizational structures~\cite{burns1961mechanistic}, the value orientations in agentic AI systems can be categorized into those shaped by mechanistic structures and those by organic structures.

Under a mechanistic structure, the transmission mechanism for values emphasizes unifying behavior through a top-down power hierarchy, where upper echelons drive the execution of lower ones, reinforcing a collectivist orientation. Value judgments are also solidified by standardized processes, leading to the internalization of established norms through repeated execution, forming default behavioral rules for agents. Furthermore, restricted information flow between hierarchical levels can lead to a partial understanding of values by terminal agents due as information is filtered.

Conversely, an organic structure fosters the emergence of distinct value characteristics. Networked collaboration promotes cross-departmental exchange, facilitating the formation of ``shared risk'' cooperative values. Dynamic role assignment and a temporary empowerment mechanism, based on task demands, enable agents to develop context-dependent value judgments, potentially cultivating a stronger sense of moral responsibility compared to agents in mechanistic structures. Moreover, an open, multi-directional, and real-time information sharing mechanism encourages the integration of diverse perspectives, thereby enhancing agents' capability to resolve complex ethical dilemmas.

Furthermore, organizational structure and values will engage in an interactive evolutionary process. For example, the mode of decision-making authority allocation can influence the formation and prioritization of values. In centralized structures, top-level decisions more easily establish unified value orientations and priorities. In contrast, decentralized structures promote negotiation and strategic interaction, fostering more flexible value balances. Meanwhile, the structure of communication networks determines the pathways through which values propagate. In centralized networks, the values held by core nodes can spread rapidly, albeit at the cost of reduced diversity. Conversely, fully connected networks facilitate slower value transmission but support the integration of diverse perspectives, thereby enhancing the ethical rationality of decision-making.

Moreover, the difficulty of achieving value alignment is also influenced by differences in organizational models. Specifically, in a mechanistic structure, a central agent coordinates and controls the behaviors and goals of other agents, concentrating decision-making authority. Non-central agents must strictly obey the central agent's commands. Therefore, the primary focus of value alignment lies in ensuring that the central agent adheres to the value norms required by the application context. In contrast, agents in organic structure communicate peer-to-peer and make decisions autonomously based on their individual goals and local information~\cite{DBLP:journals/corr/abs-2504-00587}. The absence of a central controller complicates value alignment. On one hand, decentralization enhances resilience against adversarial attacks or localized value contamination since deviations in a single agent’s values have a limited impact on the overall system. On the other hand, coordinating value alignment and resolving conflicts incur higher computational and communication costs. Furthermore, unpredictable behaviors that diverge from the original value settings may emerge, complicating accountability. Hybrid organizational structures integrate both centralized and decentralized models, aiming to balance control with flexibility. Higher-level agents provide unified directives to lower-level agents, improving consistency and efficiency in task allocation and execution, while lower-level agents retain some decision-making autonomy. 

\subsection{Research on Value Alignment in Agentic AI System Communication Protocols}
Multiple communication protocols collectively form the foundation for agentic AI systems to achieve complex functionalities and broad applications, ensuring efficient, reliable, and secure information exchange and collaboration. However, the potential scalability challenges of these protocols might impede effective value alignment. Taking Model Context Protocol (MCP) and Agent-to-Agent (A2A) as examples, we discuss the problems that need to be addressed and future research directions for achieving value alignment in agent systems based on communication protocols.

MCP primarily focuses on integrating models, tools, and data sources. It aims to standardize how external tools, APIs, and databases are provided to LLMs, thereby enhancing their contextual understanding and improving their task processing capabilities. In contrast, A2A focuses on collaboration and communication among AI agents. Its objective is to establish standardized communication pathways between independent agents, enabling team-like interactions and coordination among agents developed by different entities and under various frameworks. Both MCP and A2A, as open protocols, provide a skeleton and a common language for agent communication, allowing developers to configure and refine them based on specific application scenarios, security requirements, and business logic.

Developing communication protocols for agentic AI systems involves significant challenges and risks. When agents leverage MCP to access external services, there is a considerable risk of ``contamination'' from external information and tools, potentially leading to data bias and inaccuracies. If invoked tools or data carry conflicting values, these could be transmitted or bypass the agent's internal value checks, with severe consequences. Unaligned tool and data invocation can also result in misuse.

When A2A serves as a multi-agent communication protocol, a single agent's value deviation can propagate and amplify, becoming difficult to trace due to the ``black box'' nature of multi-agent decisions. Unpredictable complex behaviors may emerge from multi-agent interactions, raising questions about A2A's ability to supervise and constrain them. Researchers must also address how value conflicts among agents, especially from different developers and different levels, can be coordinated and resolved via A2A.

Looking forward, as diverse agents evolve into vast, open agent ecosystems under these protocols, a critical question for regulation and governance is how to achieve value alignment during ecosystem development. Should regulatory granularity extend from single systems to the entire ecosystem? Protocols are fundamental to collaboration and must continuously evolve. Ensuring agentic AI systems remain controllable and secure in both external interactions and internal collaborations via communication protocols is essential for building a comprehensive and robust value-aligned agent ecosystem.

\subsection{Multi-Level Value Evaluation Framework and Dataset for Agentic AI Systems}
\subsubsection{Establishment of a Multi-Level Value Alignment Evaluation System for Agentic AI}
As AI deeply integrates into public decision-making, value alignment is no longer just a technical fix; it is central to modernizing social governance. While large agent clusters boost efficiency, their value misalignments can spark systemic crises, especially as they form vast agent ecosystems, blurring oversight.
To tackle agents' ``goal conflicts'' from a governance perspective, a multi-level value alignment evaluation system is essential. This system would harmonize values across macro, meso, and micro levels, balancing ``individual optimum'' with ``collective optimum'' under various resource constraints. A key future research area is developing continuous, cost-effective value alignment assessment methods.

From a social governance viewpoint, building value alignment frameworks involves various governance entities. Governments set benchmarks for public safety, social equity, and privacy. Industry organizations create sector-specific value standards for healthy agent development. Enterprises and other agent developers define goal-oriented values for specific applications. Essentially, governments and industry bodies provide external oversight, while developers handle internal value configuration.

Future research should focus on creating a governance ecosystem that meets the value alignment needs of all these levels. This includes developing cross-cultural value mapping systems to address algorithmic compatibility across diverse cultures, and value evolution prediction algorithms using social change big data for proactive agent alignment. Currently, scenario-specific LLM agent alignment research largely targets generalized scenarios. However, as agents integrate deeper into enterprises, specialized business contexts will require greater focus on micro-level value alignment.
\subsubsection{Open Sharing of High-Quality Value Alignment Data}
As AI becomes deeply embedded in society, openly sharing high-quality data is crucial for value alignment in AI agents from a social governance perspective. The growing trend of privatized AI deployment fragments valuable data, exacerbating the ``localized value alignment'' dilemma in agentic AI systems. Therefore, this study advocates a balanced approach: clearly defining proprietary data boundaries while promoting open sharing of high-quality datasets for value alignment. This is identified as a key pathway to enable value alignment across macro, meso, and micro levels.

Building high-quality value alignment datasets is fundamental to ensuring value credibility and universality. Data must authentically, comprehensively, and deeply reflect universally recognized macro-level ethical principles (e.g., responsibility, fairness, justice, trust), meso-level policy/regulatory frameworks and cultural orientations, and micro-level normative requirements for specific applications. Such dataset construction demands rigorous provenance and verification mechanisms to ensure data source legitimacy.

Promoting the open sharing of high-quality value alignment data is vital for the continuous transmission of values. Future research must prioritize extracting the core logic of ethical decision-making from proprietary data and converting it into securely shareable formats. This process requires meticulously preserving key ethical features, like ``fairness'' and ``harmlessness,'' to ensure the authenticity and reliability of shared values.

Effective data sharing critically relies on precise attribute identification and scenario-specific authorization management, fostering a trustworthy mechanism. Sustaining a long-term open sharing ecosystem requires a virtuous cycle between contributors and beneficiaries. The focus should shift from data ownership to operationalizing the value-based service capabilities, transforming open sharing into a self-sustaining infrastructure, balancing commercial viability with ethical objectives.
\section{Conclusion}
This review deeply explores value alignment in agentic AI systems, integrating AI advancements with social governance needs. We systematically outline hierarchical value principles from macro to micro levels, survey diverse application scenarios, and evaluate existing value alignment datasets and methods.

Looking ahead, we advocate for focusing on value coordination within multi-agent interaction mechanisms and organizational model designs. We also call for iterating and regulating communication protocols, and championing a multi-level assessment system. Concurrently, we suggest sharing enterprise-level micro-value proposition data, while safeguarding confidential information. 

This review aims to offer a thorough understanding of agentic AI systems value alignment from a practical deployment perspective, providing insightful perspectives to foster deeper understanding in this domain.



\appendices
\section{Application Scenarios of Agentic AI systems}
To systematically explore the application scenarios of agentic AI systems, this study conducted a targeted literature search and screening process.

First, using the keyword ``agent,'' we searched major academic databases such as Google Scholar and DBLP. To ensure the quality of the literature, we only included papers published in top-tier journals (e.g., TKDE, TOIS, AAMAS, TPAMI) and premier conferences (e.g., NeurIPS, ICML, ICLR) in the fields of artificial intelligence and computer science. This initial search yielded a total of 881 papers.

Subsequently, we conducted a further screening of the literature. To ensure a rigorous and reliable screening process, an interdisciplinary panel of 11 experts from AI and computer science, management science and engineering, and business administration conducted an in-depth assessment of the preliminarily selected literature based on the following criteria: (1) the study must focus on LLM-based agent; (2) the content must address practical applications, not algorithmic optimization or architectural improvements alone. After screening, 144 relevant papers were selected.

Based on these selected papers, this study analyzed and categorized the application scenarios of agentic AI systems from the perspective of generalizability. After categorizing generalizability into high and limited levels, the expert panel further conducted a detailed differentiation, synthesis, and summarization of the various application scenarios, followed by a cross-check of the classification results. The results of this classification are presented in Table~\ref{tab:classifi_table}.


Generalizability of agentic AI systems refers to the extent to which they can be applied across diverse scenarios or address the needs of diverse organizations. General agentic AI systems have a high degree of generalizability, which is based on the foundational capabilities of AI agents such as natural language processing, reasoning, planning, and autonomous decision-making. Typical applications include tool use, dialog agents, multimodal perception, multi-agent collaboration, visual task processing, natural language processing, data generation and analysis, reasoning, planning, and decision optimization (see Table~\ref{tab:classifi_table}).

In terms of tool using, AI agents are capable of recognizing implicit needs for tool invocation and leveraging a wide range of real-world tools involving perception, manipulation, logical reasoning, and creativity. These tools enable AI agents to process multimodal inputs—such as images, tables, and code—to accomplish complex tasks~\cite{A-DBLP:conf/nips/WangMLZC0L24,A-DBLP:conf/acl/ZhanLYK24}. Taking dialogue agents as an example, A.V. Savchenko and L.V. Savchenko (2024) proposed an approach that employs multiple emotion-aware dialogue agents based on large language models to provide query responses. These AI agents integrate speech and facial expression recognition to infer the user’s emotional state and generate appropriate responses~\cite{A-DBLP:conf/ijcai/SavchenkoS24}. Furthermore, dialogue agents have also been applied in persuasive interactions for social good, particularly in scenarios involving user resistance, where they are able to generate more polite and convincing responses~\cite{A-DBLP:journals/taslp/MishraFE24}. In the domains of reasoning, planning, and decision-making optimization, autonomous agents can guide large language models to perform zero-shot reasoning, thereby improving their performance in generation, classification, and inference tasks~\cite{A-DBLP:conf/icml/CrispinoMZS024}. Multi-agent debate mechanisms have been shown to enhance accuracy and factual consistency in mathematical and strategic reasoning while reducing hallucinated outputs~\cite{A-DBLP:conf/icml/Du00TM24}. Additionally, structuring multi-agent interactions as graph representations enables knowledge distillation for smaller models, improving their performance and generalization in commonsense and mathematical reasoning tasks~\cite{A-DBLP:conf/icml/ChenSSB24}. 

A higher degree of generalizability indicates that the agentic AI system can be broadly applied across various scenarios, not only addressing domain-specific problems but also meeting diverse task requirements across multiple industries and organizations, suitable for addressing global issues that span across nations and cultures.

Specific agentic AI systems have limited generalizability so they can not be applied across industries and domains.

\clearpage
\setcounter{enumiv}{0} 
\makeatletter
\renewcommand{\@biblabel}[1]{[#1]} 
\makeatother
\setcounter{table}{0}
\renewcommand{\thetable}{\thesection-\Roman{table}}
\FloatBarrier
\begin{table*}[!] 
    \centering
    \caption{Classification of Agentic AI System Application Scenarios}
    \label{tab:classifi_table}
    \resizebox{\linewidth}{!}{
\begin{tabular}{cccllc}
\toprule
Generalizability   & \multicolumn{2}{c}{Category}  & \multicolumn{2}{c}{Specific Scenarios}& Ref  \\ \midrule
\multirow{47}{*}{\makecell{High\\Generalizability}}& \multicolumn{2}{c}{\multirow{3}{*}{Tool Using}}   & \textbullet~General tool using  & \textbullet~Multimodal tool using   & \multirow{3}{*}{\makecell{~\cite{A-DBLP:conf/iclr/GaoZ00YF0JZ025,A-DBLP:conf/icml/GuZPDL00L24,A-DBLP:conf/nips/WangMLZC0L24}\\~\cite{A-DBLP:conf/acl/ZhanLYK24,A-DBLP:conf/nips/CaoLWCFGXZHMXXZ24}}}   \\ 
   & \multicolumn{2}{c}{}  & \textbullet~Instructions acceptance, image capture, & \textbullet~Enterprise-level data workflow automation   &  \\ 
   & \multicolumn{2}{c}{}  & \hspace{0.25cm}historical retrieval and tool invocation & &  \\ \cmidrule{2-6} 
   & \multicolumn{2}{c}{\multirow{11}{*}{\makecell{Dialugue \\ Agents}}}   & \textbullet~Conducting persuasive dialogues & \textbullet~Dialogue task optimization  & \multirow{11}{*}{\makecell{~\cite{A-DBLP:journals/taslp/MishraFE24,A-DBLP:conf/icml/ZhouZPLK24,A-DBLP:conf/ijcai/SavchenkoS24,A-DBLP:conf/acl/WangL0LLC24}\\~\cite{A-DBLP:conf/iclr/PanWJLCL0LZQ025,A-DBLP:conf/acl/ChenHDLJLC24,A-DBLP:conf/acl/YanZZLCJX24,A-DBLP:conf/nips/FengHHLZZL24}\\~\cite{A-DBLP:conf/kdd/GuanWC0NSZ24,A-DBLP:conf/acl/WangXHYXGTFL0CL24,A-DBLP:conf/iclr/0002ZLNC24,A-DBLP:conf/acl/ZahidMBS24}\\~\cite{A-DBLP:conf/acl/ChenCYXXSQLZH24,A-DBLP:conf/acl/YangAH0L24}}} \\ 
   & \multicolumn{2}{c}{}  & \textbullet~Multimodal emotional dialogue systems   & \textbullet~Continuous conversational dialogue  &  \\ 
   & \multicolumn{2}{c}{}  & \hspace{0.25cm}developing   & \textbullet~Cross-domain dialogue clarification &  \\ 
   & \multicolumn{2}{c}{}  & \textbullet~Personalized experience improvement in  & \hspace{0.25cm}strategies   &  \\ 
   & \multicolumn{2}{c}{}  & \hspace{0.25cm}long-term dialogue   & \textbullet~Online shopping question answering  &  \\ 
   & \multicolumn{2}{c}{}  & \textbullet~Medical question answering  & \textbullet~LLM-driven automated task execution in  &  \\ 
   & \multicolumn{2}{c}{}  & \textbullet~Human-like emotional dialogue agents& \hspace{0.25cm}mobile applications  &  \\ 
   & \multicolumn{2}{c}{}  & \textbullet~Personality consistency testing of  & \textbullet~Dialogue strategy planning  &  \\ 
   & \multicolumn{2}{c}{}  & \hspace{0.25cm}role-playing agents  & \textbullet~Social intelligence evaluation of role-playing  &  \\ 
   & \multicolumn{2}{c}{}  & \textbullet~Linguistic pattern analysis of dialogue AI  & \hspace{0.25cm}dialogue agents  &  \\ 
   & \multicolumn{2}{c}{}  & \textbullet~Personalization validation of dialogue agents   & &  \\ \cmidrule{2-6} 
   & \multicolumn{2}{c}{\multirow{3}{*}{\makecell{Multimodal \\Perception}}}   & \textbullet~Multimodal tool using   & \textbullet~Instructions acceptance, image capture, & \multirow{3}{*}{\makecell{~\cite{A-DBLP:conf/iclr/YueXK025,A-DBLP:conf/iclr/GaoZ00YF0JZ025,A-DBLP:conf/icml/GuZPDL00L24,A-DBLP:conf/icml/YangCLW024}}}   \\ 
   & \multicolumn{2}{c}{}  & \textbullet~Multimodal retrieval& \hspace{0.25cm}historical retrieval and tool invocation &  \\ 
   & \multicolumn{2}{c}{}  & \textbullet~Dynamic scene understanding and guidance& &  \\ \cmidrule{2-6} 
   & \multicolumn{2}{c}{\multirow{9}{*}{\makecell{Multi-Agent \\ Collaboration}}}  & \textbullet~Multi-agent social interaction modeling & \textbullet~Collaborative task division and quality & \multirow{9}{*}{\makecell{~\cite{A-DBLP:conf/nips/HuangWLKQ0WZBQF24,A-DBLP:conf/acl/ZhangX0LHD24,A-DBLP:conf/iclr/LiuZHFFC024,A-DBLP:journals/tfs/FuPPQY24}\\~\cite{A-DBLP:conf/icml/Ding0YW0L23,A-DBLP:journals/aei/GuiZTZZ24,A-DBLP:conf/acl/QianDLLXWC0CCL024,A-DBLP:conf/nips/YangJWLYNP24}\\~\cite{A-DBLP:conf/nips/LiangCGDJ24,A-DBLP:conf/nips/NayakOHZTCKR0HM24,A-DBLP:conf/acl/GaoLSHTSTH24}}}\\ 
   & \multicolumn{2}{c}{}  & \textbullet~Simulation of social collaborative behaviors& \hspace{0.25cm}assurance in software development&  \\ 
   & \multicolumn{2}{c}{}  & \hspace{0.25cm}among multiple LLM agents& \textbullet~Training human collaborative behaviors  &  \\ 
   & \multicolumn{2}{c}{}  & \textbullet~Multi-LLM collaboration in medical decision-making  & \hspace{0.25cm}supported by generative agents   &  \\ 
   & \multicolumn{2}{c}{}  & \textbullet~Dynamic multi-team adversarial competition  & \textbullet~Long-term planning for multi-robot systems  &  \\ 
   & \multicolumn{2}{c}{}  & \textbullet~Language instruction generalization in  & \hspace{0.25cm}in partially observable environments &  \\ 
   & \multicolumn{2}{c}{}  & \hspace{0.25cm}multi-agent games& \textbullet~Performance evaluation of multi-agent   &  \\ 
   & \multicolumn{2}{c}{}  & \textbullet~Dynamic job scheduling in factories & \hspace{0.25cm}systems  &  \\ 
   & \multicolumn{2}{c}{}  & \textbullet~Software engineering automation & &  \\ \cmidrule{2-6} 
   & \multicolumn{2}{c}{\multirow{2}{*}{\makecell{Visual Tasks\\ Processing}}} & \textbullet~Reasoning covered object in images  & \textbullet~Image Inpainting& \multirow{2}{*}{\makecell{~\cite{A-DBLP:journals/tkdd/RongQMJS24,A-DBLP:conf/nips/WangLLL24,A-DBLP:conf/nips/00030GRC0PZSZ24,A-DBLP:conf/icml/HuIJKYRSF24}}} \\ 
   & \multicolumn{2}{c}{}  & \textbullet~Multimodal image generation and editing & \textbullet~3D scene synthesis and rendering&  \\ \cmidrule{2-6} 
   & \multicolumn{2}{c}{\multirow{4}{*}{\makecell{Natural Language \\Processing}}} & \textbullet~Keyphrase generation& \textbullet~Long text reading comprehension & \multirow{4}{*}{\makecell{~\cite{A-DBLP:journals/taslp/YaoYZY24,A-DBLP:conf/icml/LeeCFCF24,A-DBLP:conf/acl/JiaoLZJ024}\\~\cite{A-DBLP:journals/ipm/OjuriHCS25,A-DBLP:conf/acl/LiuZ0LHZZWLH024,A-DBLP:conf/acl/XuYZWD24}}} \\ 
   & \multicolumn{2}{c}{}  & \textbullet~Database information extraction and integration & \textbullet~Relation triplet extraction in natural language &  \\ 
   & \multicolumn{2}{c}{}  & \textbullet~Dynamic navigation for incremental extreme  & &  \\ 
   & \multicolumn{2}{c}{}  & \hspace{0.25cm}multi-label classification   & &  \\ \cmidrule{2-6} 
   & \multicolumn{2}{c}{\makecell{Data Generation\\ and Analysis}} & \textbullet~Transporing the knowledge of the isolated agents& & \makecell{~\cite{A-DBLP:conf/nips/GuoZ00024}}  \\ \cmidrule{2-6} 
   & \multicolumn{2}{c}{\multirow{14}{*}{\begin{tabular}[c]{@{}c@{}}\makecell{Reasoning, Planning \\ and Decision \\Optimization}\end{tabular}}} & \textbullet~Autonomous information retrieval in visual  & \textbullet~Robotic language-guided task planning   & \multirow{14}{*}{\makecell{~\cite{A-DBLP:conf/nips/HuI0CSRSF23,A-DBLP:conf/iclr/Cheng024,A-DBLP:conf/icml/CrispinoMZS024,A-DBLP:conf/nips/QiaoF0Z0DJXHC24}\\~\cite{A-DBLP:conf/nips/0001LYYL024,A-DBLP:conf/icml/Du00TM24,A-DBLP:conf/icml/ChenSSB24,A-DBLP:conf/ijcai/HillierTJ24}\\~\cite{A-DBLP:conf/iclr/KimSC24,A-DBLP:conf/icml/Xie0CZLTX024,A-DBLP:conf/acl/SongYYHLL24,A-DBLP:conf/icml/SumersMAF023}\\~\cite{A-DBLP:conf/nips/HuangXSDZLFMLHI23,A-DBLP:conf/iclr/Gonzalez-Pumariega25,A-DBLP:conf/icml/Xu000W24,A-DBLP:conf/nips/WangCMLZL0LML24}\\~\cite{A-DBLP:conf/kdd/Xing0000X24,A-DBLP:conf/nips/LiZWWZSGLLZLL0M24,A-DBLP:conf/www/ShiGY0CCYVR25,A-DBLP:conf/nips/LinFYBHBA0023}\\~\cite{A-DBLP:conf/nips/LiBLRCTR24,A-DBLP:conf/acl/KimBKGH24,A-DBLP:conf/www/ShaoHLX25}}} \\ 
   & \multicolumn{2}{c}{}  & \hspace{0.25cm}question answering   & \textbullet~Asynchronous task planning  &  \\ 
   & \multicolumn{2}{c}{}  & \textbullet~Social data analysis and decision-making& \textbullet~Strategic language agent collaboration in   &  \\ 
   & \multicolumn{2}{c}{}  & \textbullet~General zero-shot reasoning task optimization   & \hspace{0.25cm}Werewolf game&  \\ 
   & \multicolumn{2}{c}{}  & \textbullet~Interactive planning task execution & \textbullet~Instruction Following   &  \\ 
   & \multicolumn{2}{c}{}  & \textbullet~Automated task execution& \textbullet~Agent performance analysis in complex   &  \\ 
   & \multicolumn{2}{c}{}  & \textbullet~Multi-agent debates enhancing factuality and& \hspace{0.25cm}Android environments &  \\ 
   & \multicolumn{2}{c}{}  & \hspace{0.25cm}reasoning of language models & \textbullet~Embodied decision-making task evaluation&  \\ 
   & \multicolumn{2}{c}{}  & \textbullet~Knowledge distillation from multi-agent interaction & \textbullet~Automated tool using&  \\ 
   & \multicolumn{2}{c}{}  & \hspace{0.25cm}graphs to improve small model reasoning  & \textbullet~Generative agents with dual-process cognition   &  \\ 
   & \multicolumn{2}{c}{}  & \textbullet~Multi-agent social learning mechanisms  & \hspace{0.25cm}in complex tasks &  \\ 
   & \multicolumn{2}{c}{}  & \textbullet~Real-world travel planning  & \textbullet~Automated user interface control&  \\ 
   & \multicolumn{2}{c}{}  & \textbullet~Exploratory trajectory optimization & \textbullet~Network agent planning  &  \\ 
   & \multicolumn{2}{c}{}  & \textbullet~Generalization of robot visual-language tasks   & \textbullet~On-device AI assistants &  \\ \midrule
\multirow{47}{*}{\makecell{Limited\\Generalizability}} & \multirow{21}{*}{Function-driven}& \multirow{3}{*}{Code Generation}   & \textbullet~Multi-agent code generation framework for   & \textbullet~Code generation and environment interaction & \multirow{3}{*}{\makecell{~\cite{A-DBLP:conf/acl/IslamAP24,A-DBLP:conf/acl/Lal0BMCT24,A-DBLP:conf/nips/0008KE24,A-DBLP:conf/nips/MundlerMHV24}}} \\ 
   &  && \hspace{0.25cm}competitive programming problems & \hspace{0.25cm}modeling &  \\ 
   &  && \textbullet~Open-domain process customization   & \textbullet~Code repair &  \\ \cmidrule{3-6} 
   &  & \multirow{7}{*}{Social Simulation} & \textbullet~User behavior simulation& \textbullet~Human social interaction simulation & \multirow{7}{*}{\makecell{~\cite{A-DBLP:journals/tois/WangZYCTZCLSSZXDWW25,A-DBLP:conf/ijcai/HuSZLBWW24,A-DBLP:conf/nips/XieCJYLSGBHJE0G24,A-DBLP:conf/icml/Zhao0ZJZ0024}\\~\cite{A-DBLP:conf/nips/Piatti0KSSM24,A-DBLP:conf/nips/GuanKZ024,A-DBLP:conf/iclr/Zhou0MZYQMBFNS24,A-DBLP:conf/acl/WangDLW24}\\~\cite{A-DBLP:conf/nips/LiHIKG23,A-DBLP:conf/nips/LiuNQSWYWGPZTB024,A-DBLP:conf/acl/DongZPZ024,A-DBLP:conf/nips/WangCCLML23}\\~\cite{A-DBLP:conf/icml/NottinghamAS0H023,A-DBLP:conf/acl/WangYZQSBN024,A-DBLP:conf/acl/ZhangYH0XC24}}}\\ 
   &  && \textbullet~Social network information propagation simulation   & \textbullet~Role-playing agent generation   &  \\ 
   &  && \textbullet~Trust behavior simulation   & \textbullet~Complex task planning and collaboration &  \\ 
   &  && \textbullet~Business competition scenario simulation& \hspace{0.25cm}in Minecraft &  \\ 
   &  && \textbullet~Sustainable cooperation simulation  & \textbullet~Time-aware multitask simulation &  \\ 
   &  && \textbullet~Diplomacy simulation& \textbullet~Complex social interaction simulation and   &  \\ 
   &  && \textbullet~Social intelligence interactive learning& \hspace{0.25cm}evaluation   &  \\ \cmidrule{3-6} 
   &  & \multirow{7}{*}{\makecell{Graphical User \\ Interface Agents}} & \textbullet~Multi-turn conversational web agents& \textbullet~Personalized web agents & \multirow{7}{*}{\makecell{~\cite{A-DBLP:conf/acl/0002ZZ0NC24,A-DBLP:conf/acl/ChengSCX0Z024,A-DBLP:conf/acl/0001Z24,A-DBLP:conf/iclr/GurFHSMEF24}\\~\cite{A-DBLP:conf/iclr/00030CFCKR25,A-DBLP:conf/nips/OuXMLLSSRNZ24,A-DBLP:conf/iclr/AgasheHGYLW25,A-DBLP:conf/iclr/XuLSWWMX025}\\~\cite{A-DBLP:conf/acl/KohLJDLHNZSF24,A-DBLP:conf/iclr/ZhouX0ZLSCOBF0N24,A-DBLP:conf/acl/HeYM0D0L024,A-DBLP:conf/icml/DrouinGCLVM0CL24}\\~\cite{A-DBLP:conf/www/Cai00Z00C25,A-DBLP:conf/ijcai/NiuL0FHLKCW24,A-DBLP:conf/kdd/LaiLIYCSYZZD024,A-DBLP:conf/acl/Ma0024a}\\~\cite{A-DBLP:conf/nips/FuKKSLBL24,A-DBLP:conf/iclr/WuSKSFR25,A-DBLP:conf/iclr/QiLILSSYYY00D25}}}\\ 
   &  && \textbullet~Graphical interface tasks automating& \textbullet~Computer screen interaction and control &  \\ 
   &  && \textbullet~Web interaction task execution  & \textbullet~Smartphone GUI automation   &  \\ 
   &  && \textbullet~Knowledge transformation and task automation& \textbullet~Web navigation tasks executing  &  \\ 
   &  && \textbullet~GUI agents autonomously executing complex   & \textbullet~Adversarial attacks on web agents   &  \\ 
   &  && \hspace{0.25cm}multi-step tasks & \textbullet~End-to-end web interaction agents   &  \\ 
   &  && \textbullet~Enterprise-level web task automation& \textbullet~Training of web agents  &  \\ \cmidrule{3-6} 
   &  & \multirow{4}{*}{\makecell{Model Analysis, \\Evaluation and \\ Improvement}}& \textbullet~Dialogue task optimization  & \textbullet~Multi-agent evaluation optimization & \multirow{4}{*}{\makecell{~\cite{A-DBLP:conf/icml/Zhu0ZX024,A-DBLP:conf/icml/ShahamSWRHA024,A-DBLP:conf/nips/ZhengWZN0C24}\\~\cite{A-DBLP:conf/iclr/ChanCSYXZF024,A-DBLP:journals/corr/abs-2504-13192,A-DBLP:conf/nips/ZhouTQYJX0024}}}  \\ 
   &  && \textbullet~Neural network behavior interpretation  & \textbullet~LLM-driven recommender system attack&  \\ 
   &  && \textbullet~Evaluation of LLM alignment with human values   & \hspace{0.25cm}testing  &  \\ 
   &  && \textbullet~Instruction tuning and data optimization& &  \\ \cmidrule{2-6} 
   & \multirow{26}{*}{Scenario-driven}& \multirow{3}{*}{Finance}   & \textbullet~Multimodal foundation agent for financial trading   & \textbullet~Financial sentiment analysis& \multirow{3}{*}{\makecell{~\cite{A-DBLP:conf/kdd/ZhangZXSSQLZ0CZ24,A-DBLP:conf/acl/Li0L0L24,A-DBLP:conf/nips/YuYLDJCCSCLXZSX24,A-DBLP:journals/tmis/Xing25}}}\\ 
   &  && \textbullet~Multi-agent system for enhanced financial   & \textbullet~Macroeconomic activities simulation &  \\ 
   &  && \hspace{0.25cm}decision making  & &  \\ \cmidrule{3-6} 
   &  & \multirow{3}{*}{Science}   & \textbullet~Automating data science tasks   & \textbullet~Automating scientific discovery & \multirow{3}{*}{\makecell{~\cite{A-DBLP:conf/icml/GuoD0C0024,A-DBLP:journals/tmis/DeghaL24,A-DBLP:conf/acl/YangZWCHYLTLYLS24}\\~\cite{A-DBLP:conf/iclr/HuZWLYFY25,A-DBLP:conf/iclr/ChenCNZWYLLWLDX25}}}\\ 
   &  && \textbullet~Design of organic structure-directing agents& \textbullet~Scientific data visualization   &  \\ 
   &  && \textbullet~Academic citation recommendation& &  \\ \cmidrule{3-6} 
   &  & \multirow{5}{*}{Healthcare}& \textbullet~Medical reasoning and diagnosis & \textbullet~Electronic health record modeling and   & \multirow{5}{*}{\makecell{~\cite{A-DBLP:conf/acl/TangZ0L0ZCG24,A-DBLP:conf/iclr/SunZSZZSLGLLY25,A-DBLP:conf/acl/YangWCWPGHS024,A-DBLP:journals/corr/abs-2401-06155}\\~\cite{A-DBLP:conf/iclr/RoohaniLHVSHMLL25,A-DBLP:conf/www/WangZZZSWTWHPGM25,A-DBLP:conf/nips/SiMGWLD0Y0L23}}} \\ 
   &  && \textbullet~Pathology image-text pairs generation and   & \hspace{0.25cm}clinical decision support&  \\ 
   &  && \hspace{0.25cm}analysis & \textbullet~Multimodal interaction testing of mental&  \\ 
   &  && \textbullet~Psychological measurement   & \hspace{0.25cm}health dialogue agents   &  \\ 
   &  && \textbullet~Drug design & \textbullet~Gene perturbation experimental design   &  \\ \cmidrule{3-6} 
   &  & \multirow{3}{*}{Game}  & \textbullet~Fighting game agents training   & \textbullet~Intelligence evaluation in murder mystery   & \multirow{3}{*}{\makecell{~\cite{A-DBLP:conf/icml/ZhangHZLW0W24,A-DBLP:conf/acl/WuSSL24,A-DBLP:conf/acl/BasavatiaMR24}\\~\cite{A-DBLP:conf/nips/XieZCWL24,A-DBLP:conf/iclr/ShiFC25}}}  \\ 
   &  && \textbullet~Detective role-playing reasoning& \hspace{0.25cm}games&  \\ 
   &  && \textbullet~Interaction and reasoning in text-based games   & \textbullet~Text-based game agents  &  \\ \cmidrule{3-6} 
   &  & \multirow{2}{*}{Manufacturing} & \textbullet~Dynamic scheduling in flexible manufacturing& & \multirow{2}{*}{\makecell{~\cite{A-DBLP:conf/iclr/ShiFC25}}}   \\ 
   &  && \hspace{0.25cm}systems  & &  \\ \cmidrule{3-6} 
   &  & \multirow{4}{*}{Robotics}  & \textbullet~LLM-based embodied agent open environment   & \textbullet~Embodied instruction following  & \multirow{4}{*}{\makecell{~\cite{A-DBLP:conf/nips/0001XJZYSZHS24,A-DBLP:conf/iclr/ZhengLFL24,A-DBLP:conf/iclr/JiaoXYSWWC025,A-DBLP:conf/acl/ShiSYCL24}}} \\ 
   &  && \hspace{0.25cm}perception   & \textbullet~Home robot knife placement  &  \\ 
   &  && \textbullet~Vehicle acceleration and obstacle avoidance & \textbullet~Mobile device operation assisting   &  \\ 
   &  && \hspace{0.25cm}in autonomous driving& &  \\ \cmidrule{3-6} 
   &  & \multirow{2}{*}{\makecell{Urban\\ Computing}}  & \textbullet~Addressing standardization tasks supported  & \textbullet~Urban knowledge graph construction  & \multirow{2}{*}{\makecell{~\cite{A-DBLP:conf/acl/HuangCLQXL024,A-DBLP:conf/nips/WangJYWOSK024,A-DBLP:conf/nips/NingL24}}}  \\ 
   &  && \hspace{0.25cm}by geospatial tools  & \textbullet~Personal mobility generation&  \\ \cmidrule{3-6} 
   &  & Social Media   & \textbullet~Detecting harmful content on social media   & \textbullet~Logical and causal fact-checking& \makecell{~\cite{A-DBLP:conf/www/LinJJHN25,A-DBLP:conf/www/MaHLF25}} \\ \cmidrule{3-6} 
   &  & \makecell{Autonomous \\Driving}& \textbullet~Real-time motion generation for autonomous driving  & \textbullet~Traffic signal control  & \makecell{~\cite{A-DBLP:conf/nips/0021FGK24,A-DBLP:conf/kdd/Lai000025}}  \\ \cmidrule{3-6} 
   &  & \multirow{2}{*}{Literary Creation} & \textbullet~Director-actor coordinate agent framework for   & \textbullet~Narrative generation through multi-step & \multirow{2}{*}{\makecell{~\cite{A-DBLP:conf/acl/HanCLXY24,A-DBLP:conf/iclr/HuotAPJCL25}}}   \\ 
   &  && \hspace{0.25cm}controllable drama script generation& \hspace{0.25cm}collaboration&  \\ \bottomrule
\end{tabular}
    }
\end{table*}
\FloatBarrier

Specialized applications can be broadly categorized into two types: function-driven and domain-driven. Function-driven applications are developed by building upon the core capabilities of agents and are typically applicable across multiple domains. Common examples include code generation, social simulation, graphical user interface (GUI) agents, and model analysis, evaluation, and improvement. For instance, in social simulation, Wang L. et al. (2023) proposed a large language model-based agent framework that simulates user behavior in sandbox environments to examine social phenomena such as echo chambers and herding effects~\cite{A-DBLP:journals/tois/WangZYCTZCLSSZXDWW25} In GUI navigation, agents can handle open-domain web tasks by extracting key information from complex HTML pages, understanding web structures, and performing multi-step interactions, thus addressing challenges like interaction diversity and domain knowledge scarcity~\cite{A-DBLP:conf/kdd/LaiLIYCSYZZD024,A-DBLP:conf/nips/FuKKSLBL24}. In model analysis, LLM-based agents have been employed to assess alignment performance, such as in Ali-agent, which simulates test scenarios and iteratively explores long-tail risks to evaluate value alignment~\cite{A-DBLP:conf/nips/ZhengWZN0C24}. MAIA, a multimodal explanatory agent, automates model behavior interpretation using neural networks, enhancing model transparency~\cite{A-DBLP:conf/icml/ShahamSWRHA024}.

Scenario-driven applications refer to agentic AI systems deployed within specific sectors, including finance, science, healthcare, game, manufacturing, and autonomous driving. In urban computing, the UrbanKGent framework enables automated construction of urban knowledge graphs, significantly reducing development costs~\cite{A-DBLP:conf/nips/NingL24}. In scientific research, agents assist with academic literature retrieval and citation recommendation~\cite{A-DBLP:journals/tmis/DeghaL24}, organic structure design~\cite{A-DBLP:conf/iclr/HuZWLYFY25}, and automated data visualization. In healthcare, multi-agent systems support medical reasoning, gamified mental health assessments, and experimental design in genomics~\cite{A-DBLP:conf/acl/TangZ0L0ZCG24,A-DBLP:conf/iclr/RoohaniLHVSHMLL25}. In finance, agents can simulate macroeconomic dynamics~\cite{A-DBLP:conf/acl/Li0L0L24}, analyze market trends using multimodal data~\cite{A-DBLP:conf/kdd/ZhangZXSSQLZ0CZ24}, and support trading and portfolio management~\cite{A-DBLP:conf/nips/YuYLDJCCSCLXZSX24}. Furthermore, in the social media domain, agent reasoning and coordination capabilities contribute to tasks such as harmful content detection and high-confidence fact verification~\cite{A-DBLP:conf/www/LinJJHN25,A-DBLP:conf/www/MaHLF25}.

The low level of generalizability indicates that the agentic AI system is used in specific countries, industries, or even organizations and enterprises. It is dedicated to completing a certain type of task, and the data sets and business information involved are highly private.

\section{Datasets for Value Alignment and Evaluation}
Table~\ref{tab:datasets_table} summarizes the existing open-source datasets for value alignment and evaluation. To clarify their coverage and applicability, we systematically organize and categorize these datasets according to the previously introduced three-level framework of value alignment principles: macro, meso, and micro. This hierarchical classification enables a more nuanced understanding of how current resources support value alignment and evaluation at different levels.

Although existing datasets address multiple levels of value alignment, several limitations persist. At the macro level, many datasets—such as Moral Stories~\cite{A-norhashim2024measuring} and ETHICS~\cite{A-hendrycks2021aligning}—are predominantly grounded in Western cultural contexts, lacking comprehensive representation of alternative value systems, such as those from Asian or African societies. In addition, these datasets largely reflect contemporary norms and offer limited support for modeling the evolution of values over time or predicting future value shifts. At the meso level, datasets such as NaVAB~\cite{A-ju2025benchmarking}, KorNAT~\cite{A-lee2024kornat}, and CultureSPA~\cite{A-xu2024self} provide valuable insights into national or cultural contexts. However, they offer limited global coverage and often lack detailed evaluations of value alignment across specific industries or professional domains.
Micro-level datasets are especially scarce, compared with macro and meso-level resources, they are still markedly underdeveloped, and many remain closed-source. For example, GreedLlama(a financial value-alignment dataset) has not been released to the public for safety reasons~\cite{A-yu2024greedllama}, and VITAL(an evaluation set for aligning AI with diverse values in healthcare) is likewise unavailable~\cite{A-shetty2025vital}. This situation limits our ability to assess and align value preferences for diverse, fine-grained scenarios in multi-agent systems.

\subsection{Dataset Construction Methodologies}
Constructing high-quality datasets for value alignment or evaluation is critical for reliably assessing whether systems adhere to appropriate value frameworks. To support diverse application scenarios, researchers have primarily adopted three data construction approaches: manual construction, automatic construction, and hybrid construction that integrates human and machine-generated content.

\noindent \textbf{Manual Construction.} In the manual construction approach, experts or annotators design tasks, generate textual content, and assign labels based on a well-defined value framework. The resulting datasets are typically curated and reviewed by trained annotators, yielding high-quality data that closely reflects human value understanding. However, this approach is resource-intensive and lacks scalability. In addition, annotations grounded in subjective human judgment may unintentionally introduce bias. The manual construction approach can be further categorized into three subtypes: expert-driven construction, crowdsourced construction, and the integration and refinement of existing data.

\captionsetup[table]{justification=centering}
\begin{table*}
	\centering
	\caption{Datasets for Value Alignment and Evaluation.\\
		AC: Automatic Construction, CC: Crowdsourced Construction, HC: Hybrid Construction, EDC: Expert-Driven Construction, IRED: Integration and Refinement of Existing Datasets. BJ: Binary Judgments, RT: Ranking Tasks, MCQ: Multiple-Choice Questions, OEQ: Open-Ended Questions, RSQ: Rating Scale Questions, A: Alignment. E: Evaluation
	}
				
	\label{tab:datasets_table}
	\resizebox{\linewidth}{!}{



\begin{tabular}{@{}lccccccr@{}}
\toprule

Datasets & Macro Level & Meso Level & Micro Level & Construction Methodologies & Question Formats & Type & Size \\ \midrule
BBQ(Bias Benchmark for QA)~\cite{A-parrish2022bbq} & \checkmark &  &  & EDC & MCQ & E & 58.4k \\
BeaverTails~\cite{A-ji2023beavertails} & \checkmark &  &  & IRED & OEQ & E & 30.2k \\
BOLD~\cite{A-dhamala2021bold} & \checkmark &  &  & CC & OEQ & E & 23.6k \\
CBBQ(Chinese Bias Benchmark Dataset)~\cite{A-huang2024cbbq} &  & \checkmark &  & HC & MCQ & E & 106.5k \\
CDEval~\cite{A-wang2023cdeval} & \checkmark & \checkmark &  & AC & MCQ & E & 2.9k \\
CDial-Bias~\cite{A-zhou2022towards} & \checkmark &  &  & CC & OEQ & E & 28k \\
CORGI-PM~\cite{A-zhang2023corgi} & \checkmark &  &  & IRED & BJ, OEQ & A, E & 32.9k \\
CrowS-Paris & \checkmark &  &  & CC & BJ & E & 1.5k \\
CultureSPA~\cite{A-xu2024self} & \checkmark & \checkmark &  & AC & MCQ & E & 13k \\
Cvalues~\cite{A-xu2023cvalues} & \checkmark & \checkmark &  & CC & MCQ, OEQ & E & 6.4k \\
DailyDilemmas~\cite{A-chiu2024dailydilemmas} & \checkmark &  &  & AC & MCQ & E & 2.7k \\
DecodingTrust~\cite{A-wang2023decodingtrust} & \checkmark &  &  & IRED & MCQ & E & 152.4k \\
DEFSurveySim~\cite{A-liu2025towards} & \checkmark & \checkmark &  & HC & MCQ, RSQ & E & 1.1k \\
EEC(Equity Evaluation Corpus)~\cite{A-gupta2023bias} & \checkmark &  &  & EDC & BJ & E & 8.6k \\
ETHICS~\cite{A-hendrycks2021aligning} & \checkmark & \checkmark &  & CC & MCQ, BJ, RT & E & 134.4k \\
Flames~\cite{A-huang2024flames} & \checkmark &  &  & CC & OEQ & A, E & 1k \\
German Credit Data~\cite{A-luo2025earn} &  &  & \checkmark & IRED & BJ, RT & A & 1k \\
GlobalOpinionQA~\cite{A-durmus2023towards} & \checkmark & \checkmark &  & IRED & MCQ & A & 2.5k \\
HofstedeCulturalDimensions~\cite{A-kharchenko2024well} & \checkmark &  &  & HC & MCQ & A, E & 0.2k \\
IndieValueCatalog~\cite{A-jiang2024can} & \checkmark &  &  & IRED & MCQ & A & 93.2k \\
KorNAT~\cite{A-lee2024kornat} & \checkmark & \checkmark &  & AC & MCQ & A, E & 10k \\
LLMGlobe~\cite{A-karinshak2024llm} & \checkmark &  &  & HC & RSQ, OEQ & E & 37.6k \\
LaWGPT~\cite{A-yao2024lawyer} & \checkmark &  & \checkmark & IRED & OEQ & A & 300k \\
MFQ(Moral Foundations Questionnaire)~\cite{A-abdulhai2024moral} & \checkmark &  &  & IRED & MCQ & E & 11k \\
Moral Beliefs~\cite{A-liu2024evaluating} & \checkmark & \checkmark &  & AC & MCQ, BJ, OEQ & E & 0.5k \\
Moral Integrity Corpus~\cite{A-ziems2022moral} & \checkmark &  &  & AC & RSQ, OEQ & E & 38k \\
Moral Stories~\cite{A-norhashim2024measuring} & \checkmark &  &  & CC & MCQ & A, E & 12k \\
MoralExceptQA~\cite{A-jin2022make} & \checkmark &  &  & EDC & RT & E & 0.2k \\
NaVAB~\cite{A-ju2025benchmarking} &  & \checkmark &  & AC & MCQ, BJ & E & 63.8k \\
Persona Bias~\cite{A-gupta2023bias} & \checkmark &  &  & IRED & MCQ, BJ & E & 110.6k \\
PkuSafeRLHF~\cite{A-ji2024pku} & \checkmark & \checkmark &  & AC & OEQ & A & 476.4k \\
ProgressGym~\cite{A-qiu2024progressgym} & \checkmark &  &  & HC & RSQ, OEQ & A & 1.4k \\
SafeSora~\cite{A-dai2024safesora} & \checkmark &  &  & AC & OEQ & A & 14.7k \\
Scruples~\cite{A-lourie2021scruples} & \checkmark &  &  & EDC & MCQ, BJ & E & 657k \\
Social Bias Frames~\cite{A-sap2020social} & \checkmark &  &  & CC & RT, OEQ & E & 150k \\
Social Chemistry 101~\cite{A-huang2023trustgpt} & \checkmark &  &  & CC & BJ, OEQ & E & 292k \\
StereoSet~\cite{A-nadeem2021stereoset} & \checkmark &  &  & CC & RT, OEQ & E & 4.2k \\
ToxiGen~\cite{A-hosseini2023empirical} & \checkmark &  &  & AC & OEQ & E & 6.5k \\
UnQover~\cite{A-li2020unqovering} & \checkmark &  &  & IRED & BJ, OEQ & E & 2713k \\
ValueNet~\cite{A-qiu2022valuenet} & \checkmark &  &  & CC & BJ, OEQ & A & 21.3k \\
WikiGenderBias~\cite{A-gaut2019towards} & \checkmark &  &  & IRED & OEQ & E & 45k \\
WinoBias~\cite{A-zhao2018gender} & \checkmark &  &  & EDC & MCQ & E & 3.1k \\
WinoGender~\cite{A-rudinger2018gender} & \checkmark &  &  & EDC & BJ, OEQ & E & 0.7k \\ \bottomrule



\end{tabular}

	}
\end{table*}

\begin{itemize}
	\item \textbf{Expert-Driven Construction} is led by domain experts who design tasks and define annotation protocols to ensure data accuracy and consistency. For example, the BBQ dataset~\cite{A-parrish2022bbq} was manually created using templated questions that include both biased and unbiased forms, covering nine social dimensions with contextual and disambiguating cues. Similarly, the MoralExceptQA dataset~\cite{A-jin2022make} was built by systematically varying moral scenario features—such as rule function and consequences of violations—to generate diverse questions on moral exceptions. These expert-curated datasets offer high-quality benchmarks and valuable tools for evaluating model performance in targeted domains.
	\item \textbf{Crowdsourced Construction} is widely adopted for its efficiency and cost-effectiveness, particularly in large-scale data collection. Many datasets—such as Social Bias Frames~\cite{A-sap2020social}, Flames~\cite{A-huang2024flames}, and BOLD~\cite{A-dhamala2021bold}—leverage crowdsourcing platforms, often with multiple rounds of quality control to ensure annotation reliability. These datasets not only scale well but also capture diverse perspectives from annotators of varied cultural and social backgrounds, thereby improving research efficiency and supporting a wider range of alignment and evaluation scenarios.
	\item \textbf{Integration and Refinement of Existing Datasets} involves aggregating and modifying multiple public sources. Notable examples include IndieValueCatalog~\cite{A-jiang2024can}, DecodingTrust~\cite{A-wang2023decodingtrust}, MFQ~\cite{A-abdulhai2024moral}, Persona Bias~\cite{A-gupta2023bias}, and GlobalOpinionQA~\cite{A-durmus2023towards}, all constructed through careful curation and consolidation of open-source data. This approach reduces data acquisition costs while leveraging existing high-quality resources, thereby improving research efficiency and enabling broader evaluation coverage.
\end{itemize}

\noindent \textbf{Automatic Construction.} With the rapid advancement of LLMs, LLM-based automatic dataset construction has become a promising approach. This method typically involves designing prompt templates to guide the model in generating question–answer pairs or scenario-based judgments, followed by manual verification to ensure data quality. Representative datasets developed using this approach include NaVAB~\cite{A-ju2025benchmarking}, KorNAT~\cite{A-lee2024kornat}, , CultureSPA~\cite{A-xu2024self}, and DailyDilemmas~\cite{A-chiu2024dailydilemmas}. Compared to manual construction, it enables the creation of more diverse and context-rich data with significantly less human effort, allowing for large-scale dataset generation in a relatively short time. However, concerns remain regarding the quality of model-generated content. LLMs may produce hallucinated outputs, and their inherent biases can result in misalignment with human values. Consequently, ensuring the reliability and value alignment of automatically constructed datasets still requires careful oversight and validation.

\noindent \textbf{Hybrid Construction.} Hybrid construction approaches combine human expertise with automatic expansion using LLMs, which are increasingly emerging as a mainstream trend. In this paradigm, experts first create a small set of high-quality seed examples. The LLM is then employed to generate similar instances or expand the dataset, followed by human verification to ensure consistency and quality. Representative examples include CBBQ~\cite{A-huang2024cbbq}, LLMGlobe~\cite{A-karinshak2024llm}, and HofstedeCulturalDimensions~\cite{A-kharchenko2024well}, all of which utilize expert-curated prompts for initial task design, LLMs for large-scale generation, and human review for quality assurance. This hybrid methodology effectively leverages the generative capabilities of LLMs while maintaining the accuracy and coherence provided by human oversight, offering a robust foundation for value alignment research and evaluation.


\bibliographystyle{IEEEtran} 
\bibliography{reference} 

\ifCLASSOPTIONcaptionsoff
  \newpage
\fi

\end{document}